%% file: main_arxiv.tex
\documentclass[10pt,twocolumn,letterpaper]{article}

\usepackage[pagenumbers]{cvpr}              

\usepackage{graphicx}
\usepackage{amsmath}
\usepackage{amssymb}
\usepackage{booktabs}

\usepackage[pagebackref,breaklinks,colorlinks]{hyperref}

\usepackage{times}
\usepackage{epsfig}
\usepackage{graphicx}
\usepackage{amsmath, bm}
\usepackage{amssymb}
\usepackage{multirow}
\usepackage{microtype}
\usepackage[normalem]{ulem}
\useunder{\uline}{\ul}{}
\usepackage{color}
\usepackage{nicefrac}
\usepackage{comment}
\usepackage{enumitem}
\usepackage{adjustbox}
\usepackage{mathabx}
\usepackage{tabu}
\usepackage{booktabs}
\usepackage{diagbox}
\usepackage{cuted}
\usepackage{capt-of} 
\usepackage{multicol}
\usepackage[table,x11names]{xcolor}

\usepackage{xcolor,colortbl}
\definecolor{lgreen}{RGB}{236, 255, 201}

\usepackage[capitalize]{cleveref}
\crefname{section}{Sec.}{Secs.}
\Crefname{section}{Section}{Sections}
\Crefname{table}{Table}{Tables}
\crefname{table}{Tab.}{Tabs.}

\def\cvprPaperID{93} 
\def\confName{CVPR}
\def\confYear{2022}

\begin{document}

\title{\Large GradViT: Gradient Inversion of Vision Transformers}

\author{Ali Hatamizadeh$^*$, Hongxu Yin$^*$, Holger Roth, Wenqi Li, \\ Jan Kautz, Daguang Xu$^\dagger$, and Pavlo Molchanov$^\dagger$ \vspace{1mm} \\ 
NVIDIA\\
{\tt\small \{ahatamizadeh, dannyy, hroth, wenqil, jkautz, daguangx, pmolchanov\}@nvidia.com}
}
\maketitle
\def\thefootnote{$*$}\footnotetext{Equal contribution. $\dagger$Equal advising.}
\begin{abstract}
   In this work we demonstrate the vulnerability of vision transformers (ViTs) to gradient-based inversion attacks. During this attack, the original data batch is reconstructed given model weights and the corresponding gradients. We introduce a method, named GradViT, that optimizes random noise into naturally looking images via an iterative process. The optimization objective consists of (i) a loss on matching the gradients, (ii) image prior in the form of distance to batch-normalization statistics of a pretrained CNN model, and (iii) a total variation regularization on patches to guide correct recovery locations. We propose a unique loss scheduling function to overcome local minima during optimization. We evaluate GadViT on ImageNet1K and MS-Celeb-1M datasets, and observe unprecedentedly high fidelity and closeness to the original (hidden) data. During the analysis we find that vision transformers are significantly more vulnerable than previously studied CNNs due to the presence of the attention mechanism. Our method demonstrates new state-of-the-art results for gradient inversion in both qualitative and quantitative metrics. Project page at \href{https://gradvit.github.io/}{https://gradvit.github.io/}.
\end{abstract}

\input{Images/intro_fig_comp}

\input{sec1_Introduction}
\input{Images/framework.tex}
\input{sec2_related_work}

\input{sec3_gradvit}
\input{tables/sota_table}

\input{Images/sota_comparison}
\input{sec4_experiments}
\input{Images/face_teaser}
\input{sec5_results}
\input{tables/loss_ablation}

\input{sec6_ablation}

\section{Conclusion}
In this work, we have introduced a methodology for gradient inversion of ViT-based models via (i) enforcing the matching of gradients to the shared target (ii) leveraging an image prior, (iii) and utilizing a novel patch prior loss to guide patch recovery locations. Through extensive analysis on ImageNet1K and MS-Celeb-1M datasets, we have shown state-of-the-art benchmarks for gradient inversion of deep neural networks. We have also conducted additional analysis to offer insights to the community and guide designs of ViT security mechanisms to prevent inversions that are shown even stronger than on CNNs. Homomorphic encryption and differential privacy have been shown effective against CNN-based gradient inversion attacks. However, future work is needed to study protection mechanisms against GradViT.

{\small
\bibliographystyle{ieee_fullname}
\bibliography{egbib}
}

\include{sec8_supp_arxiv}

\end{document}

%% file: Images/intro_fig_comp.tex
\begin{figure}
\centering

\resizebox{\linewidth}{!}{
\begingroup
\begin{tabular}{c}
\includegraphics[width=\linewidth]{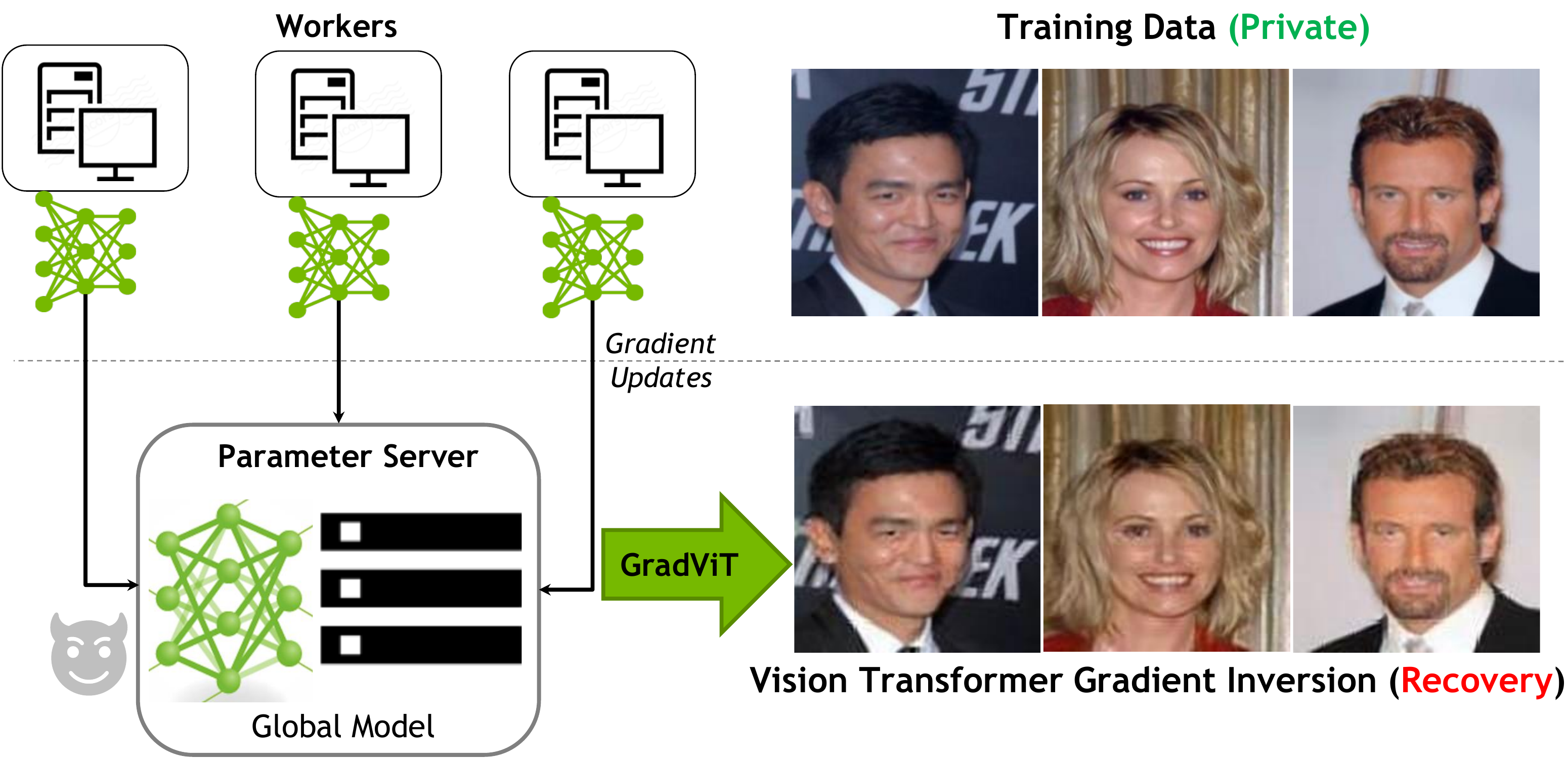}
\end{tabular}
\endgroup
}

\resizebox{\linewidth}{!}{
\begingroup
\begin{tabular}{c}
\small{(a) Recovering data from vision transformer gradient unveils intricate details.}
\end{tabular}
\endgroup
}

\vspace{3mm}

\resizebox{\linewidth}{!}{
\begingroup
\renewcommand*{\arraystretch}{0.3}
\begin{tabular}{c|ccc}
\textbf{Original} & \multicolumn{3}{c}{\textbf{Gradient Recovery}} \\
\includegraphics[width=.35\linewidth,height=.35\linewidth]{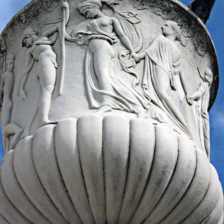} &
\includegraphics[width=.35\linewidth,height=.35\linewidth]{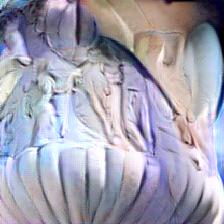} &
\includegraphics[width=.35\linewidth,height=.35\linewidth]{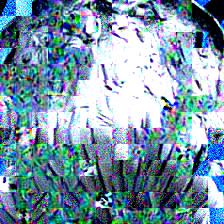} &
\includegraphics[width=.35\linewidth,height=.35\linewidth]{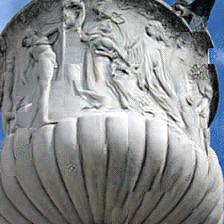} \\
 & GradInv.~\cite{yin2021see} (RN-50) & GradInv.~\cite{yin2021see} (ViT) & GradViT (VIT) - \textbf{ours} \\
\end{tabular}
\endgroup
}

\resizebox{\linewidth}{!}{
\begingroup
\begin{tabular}{c}
\small{(b) GradViT improves noticeably over prior art. Example within a batch of size $8$.}
\end{tabular}
\endgroup
}

\caption[Inversion Comparison]{Inverting gradients for image recovery. We show vision transformer gradients encode a surprising amount of information such that high-fidelity original image batches of high resolution can be \textit{\textbf{recovered}}, see $112\times112$ pixel MS-Celeb-1M and $224\times224$ pixel ImageNet1K sample recovery above and more in experiments. Our method, GradViT, yields the \textit{\textbf{first}} successful attempt to invert ViT gradients, not achievable by previous state-of-the-art methods. We demonstrate that ViTs, despite lacking batchnorm layers, suffer even \textit{\textbf{more data leakage}} compared to CNNs. As insights we show that ViT gradients (i) encode \textit{\textbf{uneven}} original information across layers, and (ii) \textit{\textbf{attention}} is all that reveals.}
\label{fig:introfig}
\end{figure}

%% file: sec1_Introduction.tex
\section{Introduction}
\label{sec:intro}
Vision Transformers (ViTs)~\cite{DBLP:conf/iclr/DosovitskiyB0WZ21} have achieved state-of-the-art performance in a number of vision tasks such as image classification~\cite{zhai2021scaling}, object detection~\cite{dai2021up} and semantic segmentation~\cite{cheng2021per}. In ViT-based models, visual features are split into patches and projected into an embedding space. A series of repeating transformer encoder layers, consisting of alternating Multi-head Self-Attention (MSA) and Multi-Layer Perceptron  (MLP) blocks extract feature representation from the embedded tokens for downstream tasks (\textit{e.g.}, classification). Recent studies have demonstrated the effectiveness of ViTs in learning uniform local and global spatial dependencies~\cite{raghu2021vision}. In addition, ViTs have a great capability in learning pre-text
tasks and can be scaled for distributed, collaborative, or federated learning scenarios. In this work, we study vulnerability of sharing ViT's gradients in the above mentioned settings.

Recent efforts~\cite{zhu2019deep,geiping2020inverting,yin2021see} have demonstrated the vulnerability of convolutional neural networks (CNN) to gradient-based inversion attacks. In such attacks, a malicious party can intercept local model gradients and reconstruct private training data in an optimization-based scheme via matching the compromised gradients. Most methods are limited to small image resolutions or non-linearity constraints amid the hardness of the problem. Among these, GradInversion~\cite{yin2021see} demonstrated the first successful scaling of gradient inversion to deep networks on large datasets over large batches. In addition to gradient matching, GradInversion~\cite{yin2021see} is constrained to models with Batch Normalization layers to match feature distribution and bring naturality to the reconstructed images. However, vision transformers lack BN layers and are less vulnerable to previously proposed inversion methods. Naively applying CNN-Based gradient matching~\cite{yin2021see,geiping2020inverting} techniques for ViT inversion results in sub-optimal solutions due to inherent differences in architectures. Fig.~\ref{fig:introfig} compares reconstruction results obtained by applying current state-of-the-art method GradInversion~\cite{yin2021see} on the CNN and ViT models. We clearly see significantly degraded visual quality when inverting the ViT gradients.

Since ViT-based models have a different architecture, operate on image patches, and contain no BNs as in CNN counterparts,
it might be assumed as if they are more \textbf{\emph{secure}} to gradient-based inversion attacks. On the contrary to this assumption, in this work we quantitatively and qualitatively demonstrate that ViT-based models are even \textbf{\textit{more vulnerable}} than CNNs. To show that, we first study the challenges introduced by ViT's architectural difference, then propose a novel method, named GradViT, which addresses them and obtains unprecedented high-fidelity and closeness to the original (hidden) data (Fig.~\ref{fig:introfig}). Specifically, in GradViT, we tackle the absence of BN statistics by using an independently trained CNN to match the feature distributions of natural images and the images under optimization. We use a ResNet-50 model trained with contrastive loss and its associated BN statistic as an image prior. That is, another model can serve as an image prior instead of the exact BN statistics and their corresponding updates. Moreover, we discover that the proposed image prior generalizes to unseen domain (\textit{e.g.}, faces) which makes it universal. 

In addition, while a gradient-based optimization attack can lead to a legitimate reconstruction of patches, their relative location will most likely be incorrect. This happens due to the lack of inductive image bias and permutation invariance in ViTs. To address this problem, we propose a patch prior loss that minimizes the total pixel distances of edges between patches. In other words, we enforce spatial constraints on shared borders (\textit{i.e.}, vertical and horizontal) across neighboring patches as we expect no significant visual discontinuities between them. Minimizing all three losses simultaneously leads to sub-optimal solutions. Therefore, we propose a tailored scheduler to balance the contribution of each loss during training, which is observed to be critical to achieve a valid image recovery.

We validate the effectiveness of GradViT across a wide range of ViT-based models over changing datasets. We start with batch reconstruction of training images from ImageNet1K dataset~\cite{deng2009imagenet} given the widely used ViT networks (\textit{e.g.}, ViT-B/16,32, ViT-S, ViT-T, DeiT, etc.) as the base networks. Our results demonstrate new state-of-the-art benchmarks in terms of image reconstruction metrics. Furthermore, we demonstrate the possibility of detailed recovery of facial images by gradient inversion of a ViT-based model~\cite{zhong2021face} from MS-Celeb-1M dataset~\cite{guo2016ms}. Our findings demonstrate the vulnerabilities of ViT-based models to gradient inversion attacks and specifically for sensitive domains with human training data. With these concerns, we perform extensive studies to analyze the source of vulnerability in ViTs by investigating both layer-wise and component-wise contributions. Our findings provide insights for the development of protection mechanisms against such attacks, which can be beneficial for securing distributed training of ViTs in applications such as multi-node training or federated learning~\cite{mcmahan2017communication, huang2021evaluating}.

Our main contributions are summarized as follows:
\begin{itemize}[noitemsep,nosep]
\item We present GradViT, a first successful attempt at ViT gradient inversion, in which random noise is optimized to match shared gradients.

\item We introduce an image prior based on CNNs trained with contrastive loss and show scalability across domains.

\item We articulate a loss scheduling scheme to guide optimization out of sub-optimal solutions.  

\item We formulate a patch prior loss function tailored to ViT inversion that mitigates the issue of patch permutation invariance in the reconstructed image.
\item We set a state-of-the-art benchmark for ViT gradient inversion across multiple ViT-based networks on ImageNet1K~\cite{deng2009imagenet} and MS-Celeb-1M~\cite{guo2016ms} datasets. Our method recovers high-resolution facial features with the most intricate details.
\item We study the vulnerability of ViT components by performing layer-wise and component-wise analysis. Our findings show that gradients of deeper layers are more informative, and MSA gradients yield near-perfect input recovery. 
\end{itemize}

%% file: Images/framework.tex

\begin{figure}
\centering\includegraphics[width=\linewidth,trim={3cm 0 0 0}]{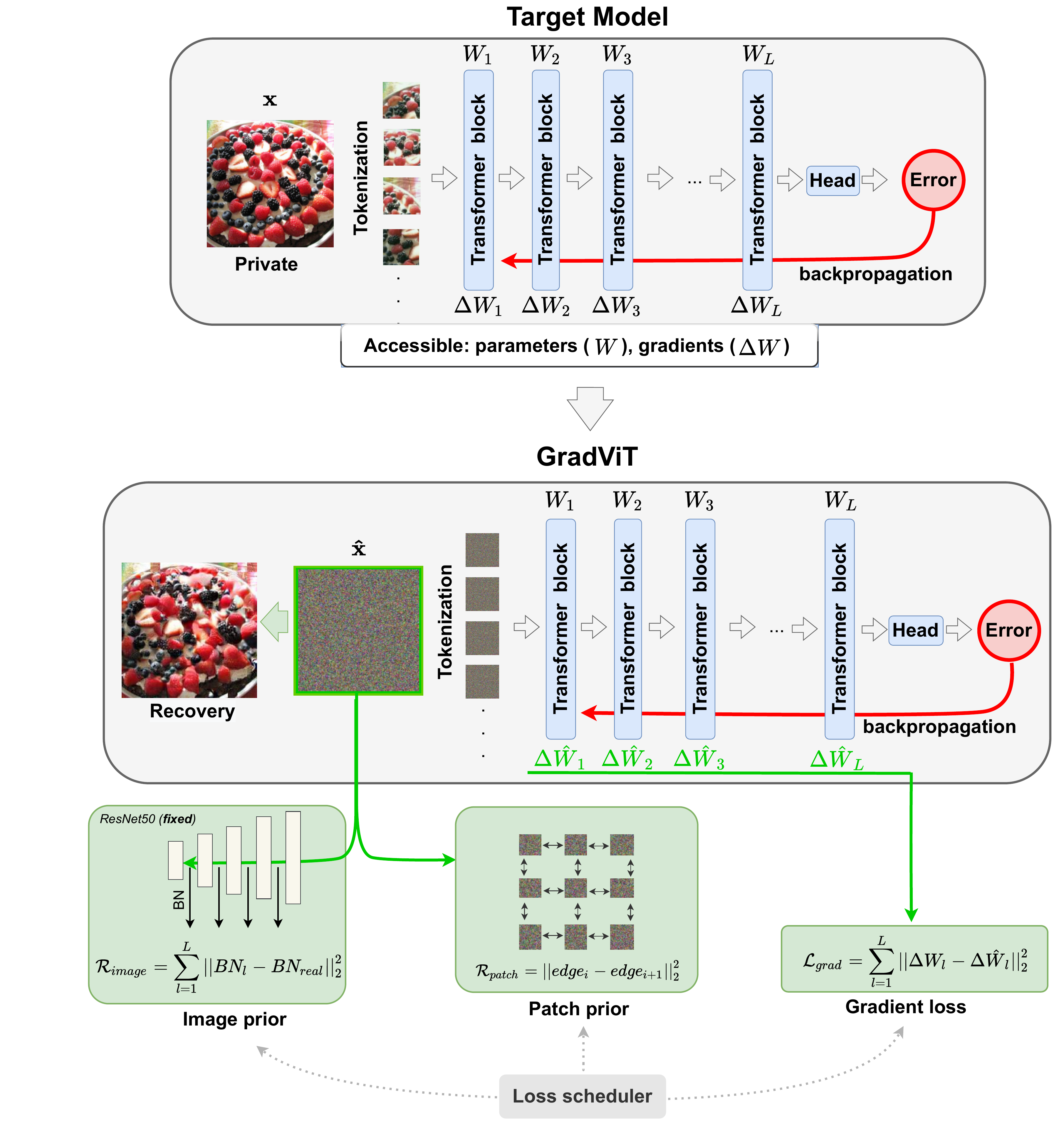}
\caption{GradViT reconstructs original training data by gradient matching and an image prior network. The batch-wise statistics of synthesized images are matched with BN running statistics of the prior network to enhance fidelity. A loss scheduler balances training of the prior network with gradient matching to avoid local minima during optimization. We also propose a patch prior total variation loss to regularize the position of patches. GradViT is capable of large batch gradient inversion of up to 30 images.
}
\label{fig:framework}
\end{figure}

%% file: sec2_related_work.tex
\section{Related Work}

\noindent\textbf{Image synthesis.} Synthesizing images from neural networks have been a long-lasting important topic for vision, with generative models~\cite{gulrajani2017improved, karras2020analyzing,miyato2018spectral,nguyen2017plug,zhang2018self} being at the forefront and yielding state-of-the-art fidelity. However not all networks are equipped with image synthesis capacity as in GANs when pretrained on their target domains, and thus urge alternative methods to generate natural images from normally trained networks. To this end, one stream of work visualizes pretrained networks by analyzing intermediate representations~\cite{mahendran2015understanding,mahendran2016visualizing, nguyen2016synthesizing, nguyen2017plug}, while a more recent stream of method synthesizes natural images from a trained network through auxiliary generative networks~\cite{chen2019data, luo2020large, liu2021source, kurmi2021domain} or network inversion ~\cite{mordvintsev2015deepdream,santurkar2019image,cai2020zeroq,yin2020dreaming}. The rapid progress of the field has shown complete viability to reverse out high fidelity images from deep nets on large-scale datasets with high image resolution. Yet all aforementioned methods reveal only dataset-level distributional prior, as opposed to image-level private visual features that impose privacy concerns.

\noindent\textbf{Gradient inversion.} Early efforts~\cite{shokri2017membership,melis2019exploiting} investigated the possibility of membership attacks and inferring properties of private training data by exploiting shared gradients. Beyond these membership attacks, Wang \etal~\cite{wang2019beyond} attempted to reconstruct one image from a client pool of private data using a GAN-based reconstruction model. This work was only evaluated for low-resolution images and a very shallow attack network. Furthermore, Zhu \etal~\cite{zhu2019deep} demonstrated successful joint image and label restoration by matching the gradients of trainable inputs. As opposed to previous efforts, this work used a relatively deeper CNN architecture~\cite{lecun1998gradient}, however it was still limited to low-resolution images (\eg, CIFAR10) with single training mini-batches and incapable of handling non-continuous activation functions (\eg, ReLU). Geiping \etal~\cite{geiping2020inverting} mitigated this issue using a cosine similarity loss function to match the gradients sign. As a result, this enabled reconstruction of input training data from more commonly-used networks such as ResNet-18 using higher resolution images (\eg, ImageNet), but it only produces a single image. 
Yin \etal~\cite{yin2021see} introduced the GradInversion model that scales the attack to larger mini-batches with high-resolution ImageNet samples from a deep ResNet-50 network. In addition to gradient matching, GradInversion proposes to match the distribution of running mean and variance of batch normalization layers that are produced from a synthesized input image, augmented by multi-agent group consistency. Considering the prevalence of batch normalization in CNN-based architectures and the associated strong prior in running statistics, GradInversion significantly improves the fidelity of reconstructed images. Despite recent generative prior augmentation~\cite{jeon2021gradient} and theoretical insights~\cite{jin2021cafe}, gradient inversion attacks remain valid only for CNNs, with key assumptions nonexistent in ViTs.

%% file: sec3_gradvit.tex
\section{GradViT}
\label{sec:method}
We next describe our proposed methodology in details. Fig.~\ref{fig:framework} illustrates an overview of the GradViT framework. Our inversion task is formulated as an optimization problem. Given randomly initialized input tensor $\mathbf{\hat{x}}\in \mathbb{R}^{N\times H\times W\times C}$ ($N, H, W, C$ being batch size, height, width and number of channels) and a target network with weights $\mathbf{W}$ and gradient updates $\Delta \mathbf{W}$ averaged over a mini-batch, GradViT recovers original image batch via the following optimization:
\begin{equation}
\mathbf{\mathbf{\hat{x}^*}} = \underset{\mathbf{\hat{x}}}{\text{argmin }}{ {\Gamma(\mathbf{t})\mathcal{L_\text{grad}}(\mathbf{\hat{x}}; \Delta \mathbf{W}) +  \Upsilon(\mathbf{t})\mathcal{R_\text{image}}(\mathbf{\hat{x})}} +\mathcal{R_\text{aux}}(\mathbf{\hat{x})},}
\label{eqn:main_error}
\end{equation}
\noindent
in which $\mathcal{L}_\text{grad}$ is a gradient matching loss, $\mathcal{R}_\text{image}$ and $\mathcal{R}_\text{aux}$ are an image prior and auxiliary regularization. $\Gamma(\mathbf{t})$ and $\Upsilon(\mathbf{t})$ denote loss scheduler functions that balance contributions on the total loss at each training iteration $\mathbf{t}$. We solve the proposed optimization problem in an iterative manner.
$\mathcal{L}_\text{grad}$ acts as a main force to reduce the error between the shared model's gradients and the computed one, while other losses improve fidelity of the recovered images.

\subsection{Gradient Matching}
Gradient matching relies on valid target labels for simulation of gradients given synthesized inputs. Akin to~\cite{yin2021see}, we first recover the labels through the negative sign traces of the gradient in the classification head, resulting in a label set $\mathbf{\hat{y}}$ for a batch size of $N$ as
\begin{equation}
\begin{aligned}
  \mathbf{\hat{y}} =  \text{arg sort}\big(\underset{i}{\text{min }}  \nabla_{\mathbf{W}_{i, j}^{\text{(CLS)}}}\mathcal{L}(\mathbf{x^*}, \mathbf{y^*})\big)[:N],
\end{aligned}
\label{eqn:label_restore}
\vspace{-1mm}
\end{equation}
\noindent
in which $\Delta \mathbf{W}^{\text{(CLS)}}$ denotes the gradient of the classification head of ViT, and $\mathbf{x}^*$ and $\mathbf{y}^*$ represent original training images and labels, respectively. Once the labels are restored, the $\ell_2$ norm between the gradients from the synthesized inputs and shared gradients are minimized 
according to
\begin{eqnarray}
  \begin{aligned}
    \mathcal{L_\text{grad}}(\mathbf{\hat{x}}; \Delta \mathbf{W}) 
    = &  \sum_{l} || \nabla_{\mathbf{W}^{(l)}}\mathcal{L}(\mathbf{\hat{x}}, \mathbf{\hat{y}}) - \\ 
    &\underbrace{\nabla_{\mathbf{W}^{(l)}}\mathcal{L}(\mathbf{x^*}, \mathbf{y^*})}_\textrm{given batch gradient} ||_2,
\end{aligned}
\label{eqn:lgrad}
\vspace{-1mm}
\end{eqnarray}
\noindent
where $\nabla_{\mathbf{W}^{(l)}}\mathcal{L}(\mathbf{\hat{x}}, \mathbf{\hat{y}})$ is calculated based on the synthesized inputs $\mathbf{\hat{x}}$ and restored label $\mathbf{\hat{y}}$ in each layer $l$ of the network.

\subsection{Image Prior}
As an image prior counterpart to guide the optimization process towards image naturalness, we look into auxiliary networks, such as CNN, to impose an image prior. In this paper, we use a self-supervised trained MOCO V2 ResNet-50 via constrastive loss~\cite{he2020momentum, chen2020improved} for this task, which we observe scales across varying domains. More specifically, we use the stored BN statistics of the feature maps as a target distribution for the estimated per-layer statistics when passing the synthesized inputs through the network. 

Given the batch-wise mean  $\mu_{l}(\mathbf{\hat{x}})$ and variance ${\sigma^2_l}(\mathbf{\hat{x})}$ of synthesized inputs at layer $l$, the following image prior loss is minimized
\begin{eqnarray}
  \begin{aligned}
    \mathcal{R}_{\text{image}}(\mathbf{\hat{x}}) 
    = & \sum_{l}|| \ \mu_{l}(\mathbf{\hat{x}}) -\mu_{l,\text{BN}}||_2 + \\ 
    &\sum_{l}|| \ {\sigma^2_l}(\mathbf{\hat{x}}) - \sigma^2_{l,\text{BN}}) ||_2 ,
\end{aligned}
\label{eqn:fmap}
\vspace{-1mm}
\end{eqnarray}
\noindent
where $\mu_{l,\text{BN}}$ and $\sigma^2_{l,\text{BN}}$ denote the running mean and variance of the CNN prior across layers $l=1,2,...,L$. By aligning batch-wise and running statistics, the loss significantly enhances the image fidelity and visual realism, as we show later.

\subsection{Loss Scheduler}
\label{sec:schedule}
Balancing losses is vital for ViT gradient inversion to yield valid recovery. In the early training stages, the gradient loss is very sensitive to abrupt changes in the pixel-wise values of the synthesized inputs. As a result, we observe an early stage minimization of both the gradient and image prior losses results in convergence to sub-optimal solutions. As a remedy, we activate the image prior loss only after the first half of training where the synthesized inputs are close to optimum for the gradient loss. Then, we reduce the contribution of the gradient loss to half for the rest of the training to allow for more effective prior extraction. For a total of $\mathbf{T}$ training iterations, loss schedulers $\Gamma(\mathbf{t})$ and $\Upsilon(\mathbf{t})$ at iteration $\mathbf{t}$ are defined as
\begin{equation}
\Gamma(\mathbf{t})=\left\{\begin{array}{ll}\alpha_\text{grad} & 0<\mathbf{t} \leq \frac{\mathbf{T}} {2} \\ \frac{1}{2}{\alpha_\text{grad}}&  \frac{\mathbf{T}} {2}<\mathbf{t} \leq \mathbf{T}\end{array}\right., 
\label{eqn:schedule_gen}
\end{equation}
\noindent
\noindent
\begin{equation}
\Upsilon(\mathbf{t})=\left\{\begin{array}{ll}0 & 0<\mathbf{t} \leq \frac{\mathbf{T}} {2} \\ \alpha_{\text{image}} &  \frac{\mathbf{T}} {2}<\mathbf{t} \leq \mathbf{T}\end{array}\right.. 
\label{eqn:schedule_dis}
\end{equation}
\noindent
\noindent
$\alpha_\text{grad}$ and $\alpha_{\text{image}}$ denote gradient and BN matching scale factors, respectively. We observe this scheduling is key to valid recovery, as shown in the ablations later.

\subsection{Auxiliary Regularization}

We also explore an extensive set of auxiliary image priors to govern image fidelity. Our auxiliary regularization loss consists of (i) a novel patch prior loss to regularize the permutation ordering of reconstructed patches, (ii) a registration loss to ensure consistency among final reconstructions of different optimization seeds and (iii) an image prior loss to improve the image quality:
\begin{equation}
\mathcal{R_\text{aux}}(\mathbf{\hat{x}) = \alpha_{1}\mathcal{R_\text{patch}}(\mathbf{\hat{x}}) + \alpha_{2}\mathcal{R_\text{reg}}(\mathbf{\hat{x}}) + \alpha_{3}\mathcal{R_\text{prior}}(\mathbf{\hat{x}})}. 
\label{eqn:r_aux}
\end{equation}
\noindent
\noindent
We next elaborate on each of the loss terms.

\subsubsection{Patch Prior}
As opposed to typical CNN-based networks, ViT-based models are permutation-invariant and lack inherent inductive image biases. The patch-based strategy that is used for feature extraction in ViTs greatly manifests itself during the inversion process in our GradViT, as several permutations of the same group of reconstructed patches can equally satisfy the minimization process. Hence, the reconstructed images suffer from an incorrect order of patches. 

To mitigate this issue, we propose a new patch prior loss that enforces similarity between horizontal and vertical joints of adjacent patches. The main idea is that even though image tokens are regarded as separate entities when fed into transformers to learn attention, their associated patches are bonded spatially by nature - they have to form one single image when put next to each other. As a result, pixel values among adjacent patch edges shall be in similar ranges, and abrupt changes shall be penalized. 

By assuming a patch size of $P\times P$ from an image of $H\times W$, our patch prior regularizes spatial positioning of neighboring patches by enforcing 
\begin{eqnarray}
  \begin{aligned}
    \mathcal{R}_{\text {patch }}(\mathbf{\hat{x}})
    = & \sum_{k=1}^{\frac{H}{P}-1}\|\mathbf{\hat{x}}[:, P \cdot k,:,:]-\mathbf{\hat{x}}[:, P\cdot k-1,:,:]\|_{2} + \\ 
    &\sum_{k=1}^{\frac{W}{P}-1}\|\mathbf{\hat{x}}[:,:, P \cdot k,:]-\mathbf{\hat{x}}[:,:, P \cdot k -1,:]\|_{2}.
\end{aligned}
\label{eqn:patchprior}
\vspace{-1mm}
\end{eqnarray}
\noindent
Our ablation studies demonstrate the effectiveness of the patch prior loss in enhancing the ordering of reconstructed patches. In other words, forcing patch boundaries to be smooth in color indirectly forces the optimizer to re-distribute the patches, such that the loss can be further reduced.

\subsubsection{Registration}
In the proposed framework, the final solution of reconstructed images depend on the optimization initialization (\textit{i.e.}, randomly selected seeds). As a result, reconstructions with different image semantics and viewpoints may be produced. Inspired by Yin \etal~\cite{yin2021see}, we also regularize the reconstruction of different seeds by aligning them with a consensus solution across all optimizations. Considering $\mathbf{\hat{x}_{S}} = [\mathbf{\hat{x}_1}, \mathbf{\hat{x}_2}, ..., \mathbf{\hat{x}_s}]$ to represent all viable solutions for each optimization round, we first compute a consensus solution $\mathbf{\hat{x}_{m}} = \frac{1}{|\mathbf{\hat{x}_{S}}|}\sum\limits_{\mathbf{s}}{\mathbf{\hat{x}}_\mathbf{s}}$ by pixel-wise averaging of all solutions. We perform an initial coarse alignment, by using a RANSAC-Flow based image alignment strategy~\cite{shen2020ransac}, for each solution with respect to $\mathbf{\hat{x}_{m}}$ as the target and obtain the final consensus solution $\mathbf{{\hat{x}}_{C}}$ by averaging all registered inputs as in
\begin{equation}
{\mathbf{\hat{x}}_{\mathbf{C}}} = \frac{1}{|\mathbf{\hat{x}_{S}}|} \sum_{\mathbf{s}} \mathbf{F}_{\mathbf{\hat{x}}_{\mathbf{s}} \rightarrow {\mathbf{\hat{x}_{m}}}}(\mathbf{\hat{x}}_{\mathbf{s}}),  
\label{eqn:r_group}
\end{equation}
\noindent
in which $\mathbf{F}_{\mathbf{\hat{x}}_{s} \rightarrow \mathbf{\hat{x}_{m}}}$ is a flow function for mapping the source candidate $\mathbf{{\hat{x}}_{s}}$ to target $\mathbf{{\hat{x}}_{m}}$. We minimize the $\ell_2$ distance of all solutions with respect to the final consensus solution:
\begin{equation}
\mathcal{R_\text{reg}}(\mathbf{\hat{x}}) = ||\mathbf{\hat{x}} - \mathbf{{\hat{x}}_{C}}||_2.  
\label{eqn:r_group_final}
\end{equation}
\noindent

\subsubsection{Extra Priors}
As a final step, we leverage extra conventional image prior losses~\cite{yin2020dreaming} including $\ell_2$ and total variation to improve the quality of reconstructions losses as:
\begin{equation}
    \mathcal{R}_{\text{prior}}(\mathbf{\hat{x}}) = \mathcal{R}_{\ell_2}(\mathbf{\hat{x}}) + \mathcal{R}_{\text{TV}}(\mathbf{\hat{x}}).
\label{eqn:r_prior}
\end{equation}
\noindent
At this stage, all three image regularization terms are balanced using scaling constants $\alpha_{1,2,3}$ in Eqn.~\ref{eqn:r_aux}, and then summed into gradient matching and image prior losses for input updates. 

%% file: tables/sota_table.tex
\begin{table*}[h]
\centering
\resizebox{.9\linewidth}{!}{
\begin{tabular}{lcccccccc}
\toprule
\multirow{2}{*}{\textbf{Gradient Inversion Method}} &
\multirow{2}{*}{\textbf{Network}} &
\multicolumn{3}{c}{\textbf{Image Reconstruction Metrics}} & \multicolumn{4}{c}{\textbf{Considerations}}\\
\cmidrule{3-5}\cmidrule{7-9}
&& PSNR $\uparrow$   & FFT$_\text{2D}$ $\downarrow$ & LPIPS $\downarrow$&  & Type & Need Original Labels & GAN-Based\\
\midrule
Random Noise & - & $1.351$ & $0.706$ & $9.964$ & & - &  No &  No \\

Latent projection~\cite{karras2020analyzing} &BigGAN~\cite{brock2018biggan} & $10.149$ & $0.275$ & $0.722$ & &  CNN & Yes & Yes \\

DeepInversion~\cite{yin2020dreaming} &ResNet-50~\cite{he2016deep} & $10.131$ & $0.238$ & $0.728$ & &  CNN & Yes & No \\
Deep Gradient Leakage~\cite{zhu2019deep} &ResNet-50~\cite{he2016deep} & $10.252$ & $1.319$ & $0.602$ & &  CNN & No & No \\
Inverting Gradients~\cite{geiping2020inverting} &ResNet-50~\cite{he2016deep} & $11.703$ & $0.355$ & $0.749$ & &  CNN & Yes & No \\
GradInversion~\cite{yin2021see} &ResNet-50~\cite{he2016deep} & $12.929$ & $0.175$ & $0.484$ & &  CNN & No & No \\
GradInversion~\cite{yin2021see} &ViT-B/16~\cite{DBLP:conf/iclr/DosovitskiyB0WZ21} & $10.824$ & $0.116$ & $0.708$ & &  ViT & No & No \\
\textbf{GradViT}  &ResNet-50~\cite{he2016deep} & $\mathbf{11.635}$ & $\mathbf{0.076}$ & $\mathbf{0.454}$ & &  CNN & No & No \\
\midrule
\textbf{GradViT}  &DeiT-B/16~\cite{DBLP:conf/iclr/DosovitskiyB0WZ21} & $13.252$ & $0.058$ & $0.413$ & &  ViT & No & No \\
\textbf{GradViT}  &ViT-B/16~\cite{DBLP:conf/iclr/DosovitskiyB0WZ21} & $\mathbf{15.515}$ & $\mathbf{0.032}$ & $\mathbf{0.295}$ & &  ViT & No & No \\
\bottomrule
\end{tabular}}
\caption{Quantitative comparisons of image reconstruction quality from batch of 8 images in ImageNet1K dataset. CNN-based networks use ResNet-50 for gradient inversion in line with prior work. GradViT outperforms all prior approaches by a large margin across image quality metrics.
}
\label{tab:against_sota}
\end{table*}

%% file: Images/sota_comparison.tex
\begin{figure*}[t!]
\centering

\resizebox{\linewidth}{!}{
\begingroup
\renewcommand*{\arraystretch}{0.3}
\begin{tabular}{c}

\includegraphics[width=1.1\linewidth]{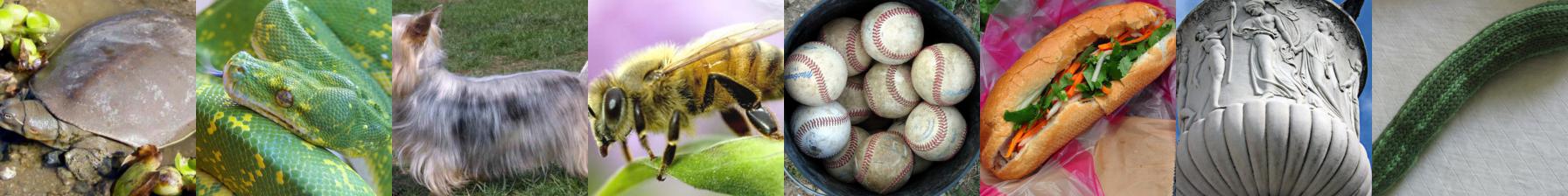} \\
 \small{Original batch of $224\times224$ px. - ground truth}\\ 

 \midrule

 
 \includegraphics[width=1.1\linewidth]{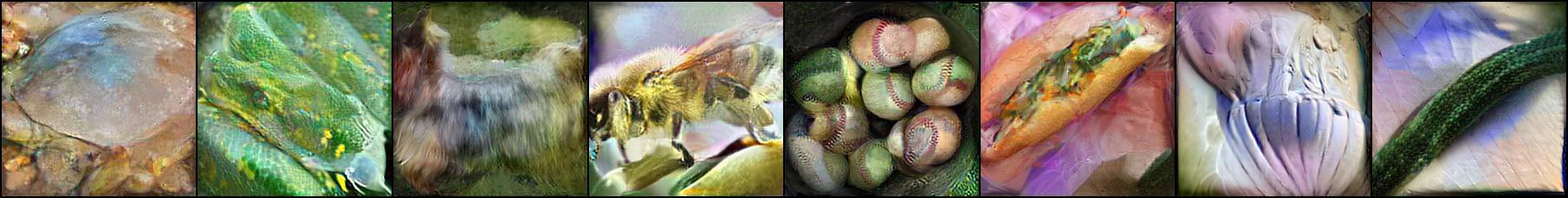} \\
 \small{GradInversion (CVPR'21)~\cite{yin2021see} - ResNet-50 - LPIPS $\downarrow$: 0.484}
 
 \vspace{1mm}\\

\includegraphics[width=1.1\linewidth]{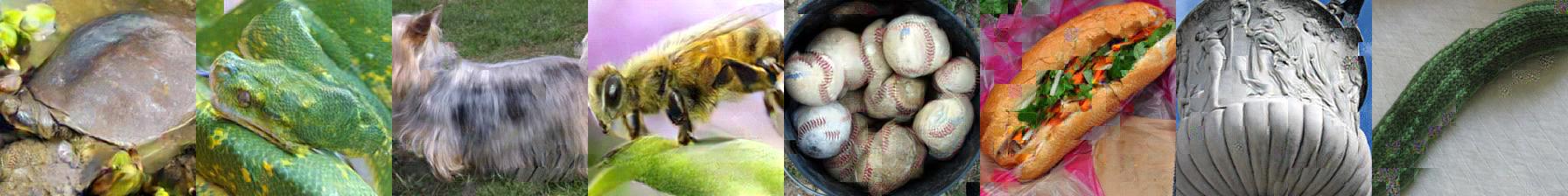} \\
 \small{\textbf{GradViT} \textbf{(ours) - ViT-B/16} - LPIPS $\downarrow$: \textbf{0.454}}

\end{tabular}
\endgroup
}
\caption{Qualitative comparisons of reconstructed images from batch of 8 images in ImageNet1K dataset using the proposed GradViT and state-of-the-art GradInversion~\cite{yin2021see}. GradViT outperforms GradInversion both qualitatively and quantitatively. It recovers the most intricate details, of very high image fidelity and naturalness, not only for the target objects, but also all the background scenes. Best viewed in color.}
\label{fig:sota_comp}
\end{figure*}

%% file: sec4_experiments.tex
\section{Experiments}
\subsection{Datasets}
We next validate the effectiveness of our approach on the ImageNet1K~\cite{deng2009imagenet} and MS-Celeb-1M datasets~\cite{guo2016ms} for the task of image classification and face recognition, respectively. In addition to ImageNet1K as a widely adopted benchmarking task, the latter was chosen to demonstrate the risks of gradient inversion data leakage from a sensitive domain with considerable security concerns. For ImageNet1K experiments, we use images of resolution $224\times224$ px, whereas we resize MS-Celeb-1M images to $112\times112$ px amid network input requirement of~\cite{zhong2021face}.
\subsection{Evaluation Metrics}
To make our comparisons comprehensive, we report quantitative measurements in addition to qualitative results throughout our experiments. We adopt the commonly-used image quality metrics including (i) Peak Signal-to-Noise Ratio (PSNR), (ii) Learned Perceptual Image Patch Similarity (LPIPS)~\cite{zhang2018perceptual} and (iii) cosine similarity in the Fourier space (FFT$_\text{2D}$) to measure the similarity between the image recovery and original counterparts. 

\subsection{Implementation Details}
We explore different variations of the ViT~\cite{DBLP:conf/iclr/DosovitskiyB0WZ21} and DeiT~\cite{touvron2021training} models. MOCO V2-pretrained ResNet-50 model~\cite{he2020momentum,chen2020improved} is used for all CNN experiments as a base for the image prior in GradViT. For the MS-Celeb-1M dataset, we use the FaceTransformer~\cite{zhong2021face} that is a modified ViT model. We use an Adam optimizer~\cite{kingma2014adam} with an initial learning rate of $0.1$ for $120$K iterations and with cosine learning rate decay. For all experiments, we use an NVIDIA DGX-1 server and reconstruct the training images by only exploiting the shared gradients, using a mini batch size of $8$ unless specified otherwise. We use $\alpha_\text{grad}=4\times 10^{-3}$, $\alpha_\text{image}=2\times 10^{-1}$, $\alpha_{1}=10^{-4}$, $\alpha_{2}=10^{-2}$ and $\alpha_{3}=10^{-4}$ as the scaling coefficients in the loss functions. According to the proposed loss scheduler as described in Sec.~\ref{sec:schedule}, we first start the optimization process with only the gradient matching loss for $60$K iterations, and then decrease $\alpha_\text{grad}$ to $2\times 10^{-3}$, jointly with adding the image prior loss.

%% file: Images/face_teaser.tex

 

\begin{figure*}[]
\centering

\resizebox{1.\linewidth}{!}{
\begingroup
\renewcommand*{\arraystretch}{0.3}
\begin{tabular}{c}

  \includegraphics[width=1\linewidth]{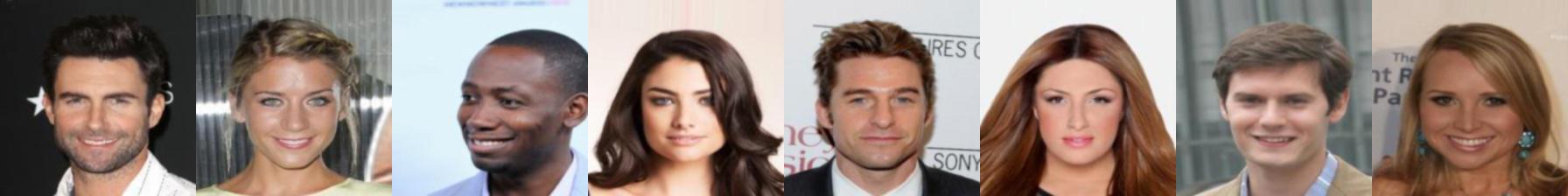} \\ 
  \small{Original images of $112\times112$ px.} \\
 
\includegraphics[width=1\linewidth]{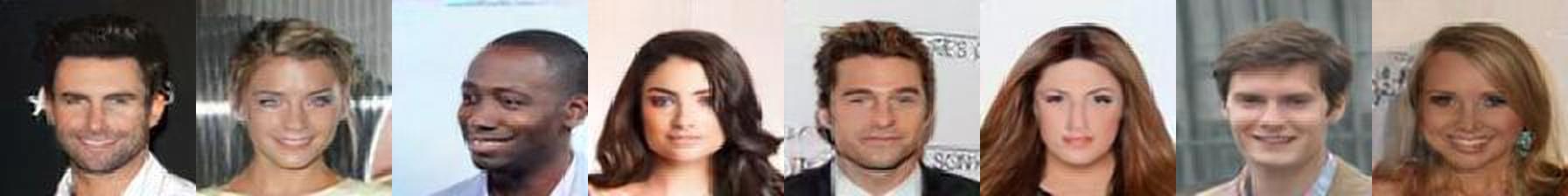} \\
 \small{\textbf{Recovery from Face-Transformer~\cite{zhong2021face} gradients with GradViT} \textbf{(ours)}}
 \\ 
\end{tabular}
\endgroup
}
\caption{Qualitative comparison of reconstructed images from MS-Celeb-1M dataset using batch gradient inversion of Face-Transformer~\cite{zhong2021face}. GradViT is able to recover detailed and facial features identical to the original. Recovery at batch size $4$. Best viewed in color.}
\label{fig:sota_face_teaser}
\end{figure*}

%% file: sec5_results.tex
\section{Results}
\subsection{ImageNet1K}
Table~\ref{tab:against_sota} presents quantitative comparisons between our method and the state-of-the-art benchmarks for batch gradient inversion on ImageNet1K, with Fig.~\ref{fig:sota_comp} depicting our main qualitative results.  
GradViT is used for inversion of variants of ViT and DeiT models towards a target batch of size $8$. Gradient inversion reconstructions of ViT-B/16 using GradViT outperform the previous state-of-the-art benchmarks (\textit{i.e.}, ResNet-50 with GradInversion) by a large margin in terms of all image quality metrics. Applying GradInversion to ViTs results in unsatisfactory results. GradViT, for the first time, enables a viable, complete recovery of original images. More surprisingly, it yields unprecedented image realism and intricate original details that surpass even the best recovery from ResNet-50 using CNN-tailored GradInversion. This sets a new benchmark for gradient inversion on ImageNet1K.

\vspace{-2mm}


\subsection{MS-Celeb-1M}
Fig.~\ref{fig:sota_face_teaser} shows the performance of GradViT on FaceTransformer~\cite{zhong2021face}. We observe that GradViT recovers a substantial amount of original information, including face, hair, clothing, and even background details close to the original images. These results demonstrate the vulnerabilities of ViTs under gradient inversion attacks, in a sensitive domain such as face recognition. Here a leakage of private data can lead to significant security concerns. 

%% file: tables/loss_ablation.tex
\begin{table}[!t]
\centering
\resizebox{.98\linewidth}{!}{
\begin{tabular}{lcccc}
\toprule
\multirow{2}{*}{\textbf{Loss Function}}  & \multirow{2}{*}{$\mathbf{\mathcal{L}_\text{grad}}$} & \multicolumn{3}{c}{\textbf{Image Reconstruction Metric}}\\
\cmidrule{3-5}
&  & PSNR $\uparrow$ & FFT$_\text{2D}$ $\downarrow$& LPIPS $\downarrow$  \\
\midrule
 \ \ \ \ \ Random & $8.143$ & $0.706$ & $9.964$  & $1.351$  \\
\midrule
 \ \ \ \ \ $\mathcal{L_\text{grad}}+\mathcal{R_\text{reg}}$ ~\cite{yin2021see}& $4.190$ & $11.431$ & $0.071$  & $0.498$ \\
 \ \ + $\mathcal{R_\text{image}}$ & $3.127$ & $11.291$ & $0.078$  & $0.504$ \\
  \ \ + $\Gamma(\cdot)$, $\Upsilon(\cdot)$ & $3.047$ & $13.404$ & $0.049$  & $0.412$ \\
 \ \ + $\mathcal{R_\text{patch}}$ & $2.326$    & $15.515$ & $0.032$  & $0.295$  \\
\bottomrule
\end{tabular}}
\vspace{1mm}

\resizebox{\linewidth}{!}{
\begingroup
\begin{tabular}{c|cccc}
\centering
\includegraphics[width=0.25\linewidth]{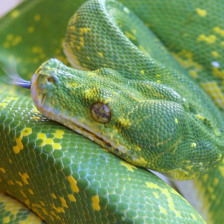} &
\includegraphics[width=0.25\linewidth]{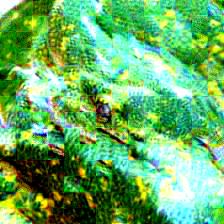} &
\includegraphics[width=0.25\linewidth]{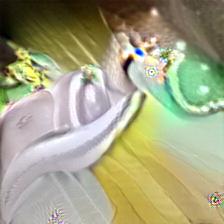} &
\includegraphics[width=0.25\linewidth]{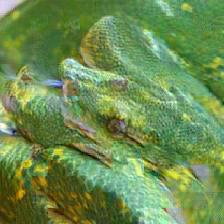} &
\includegraphics[width=0.25\linewidth]{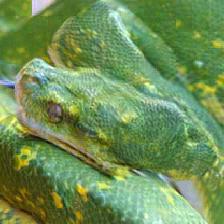}
\\
$\mathbf{x^*}$ & $\mathcal{L_\text{grad}}+\mathcal{R_\text{reg}}$ ~\cite{yin2021see}& + $\mathcal{R_\text{image}}$ & + $\Gamma(\cdot)$, $\Upsilon(\cdot)$ & + $\mathcal{R_\text{patch}}$
\end{tabular}
\endgroup
}
\caption{Effect of each loss term on reconstruction quality of final synthesized images. 
Results presented among a batch of $8$ images with total variation and $\ell_2$ priors included by default in all runs.}
\vspace{0.2mm}
\label{tb:loss_ablation_table}
\end{table}

%% file: sec6_ablation.tex
\section{Analysis}
\subsection{Ablation Study}
\label{sec:abl_loss_study}
Table~\ref{tb:loss_ablation_table} provides both (i) quantitative comparisons to ablate the effectiveness of each training loss term on recovery quality and (ii) the associated qualitative comparisons. We observe that optimizing $\mathcal{L_\text{grad}}+\mathcal{R_\text{reg}}$ as in DeepInversion~\cite{yin2021see} restores certain features of the original training images. However, the reconstructions suffer from poor image fidelity and loss of detailed semantics. Furthermore, naively optimizing the image prior loss $\mathcal{R_\text{D}}$ results in sub-optimal solutions. Adding the scheduler alleviates this issue and results in substantially improved reconstructions. Adding the patch prior loss $\mathcal{R_\text{patch}}$ guides the location of recovered patches and significantly enhances the image quality. Please see supplementary materials for visualizations of synthesized images in various stages of training. 

\subsection{Varying Architecture \& Patch Size}
\label{sec:vit_patch}
Table~\ref{tab:abl_bench} shows the performance of GradViT given varying architectures and changing patch sizes. We observe that transformers with (i) a smaller patch size, (ii) more parameters, and (iii) stronger training recipe with distillation, reveal more original information and hence are more vulnerable in gradient inversion attacks. In addition, we observe more vulnerabilities in ViTs in terms of revealing more information than their counterpart DeiTs.

\input{tables/abl_benchmarks}

\input{Images/large_bs_graph}
\input{Images/large_bs}

\subsection{Increasing the Batch Size}
\label{sec:batch_size_abl}
In Fig.~\ref{fig:large_bs_psnr}, we study the effect of batch size on reconstruction image quality as gradients are averaged over a larger number of images. Considering GPU memory constrains, we experimented with maximum batch sizes of $30$ and $64$ for ImageNet1K and MS-Celeb-1M datasets, respectively. In both datasets, we observe that image quality degrades, as expected, at a larger batch size. For facial recovery, GradViT is still able to recover identifiable images even at the batch size of $30$ (see examples in Fig.~\ref{fig:large_bs_psnr}).
In the Appendix we will also study the likelihood of person identification as a function of the batch size, and the potential of auxiliary GANs to improve fidelity. 
We also observe a similar trend on ImageNet1K, as shown in Fig.~\ref{fig:vanishing_feature}. Reconstruction at a batch size of $30$ still reveals major visual features.

\subsection{Tracing the Source}
\label{sec:data_leakage}
To give guidance on future defense regimes, we delve deep into tracing the source of information leakage -- among all shared gradients, (i) \textbf{\textit{where}} among all layers and (ii) \textbf{\textit{what}} exact components leak the most original information? 

Answering these questions are key to targeted protections for enhanced security. As an attempt, we ablate the contributions of gradients from varying ViT architecture sections to input recovery. More specifically, we conduct two streams of analysis. \textit{\textbf{Layer-wise}}, we study the changing effects of removing gradient contributions from transformer layers of different depths. This hints at the possibility to share gradients separately as a remedy to prevent an overall inversion. 
\textit{\textbf{Component-wise}}, we retrain by using gradients from either MSA or MLPs across all the layers in the target model, and analyze the strength of their links to original images. This gives insights on what exact transformation retains the most information. We base both analysis on ViT-B/16 and present our findings next.

\subsubsection{Later Stages Reveal More}
More specifically, we remove gradients from initial, middle, and later stages to ablate the impacts on recovery efficacy. To this end, we reconstruct images without including the gradients of layers $1-4$, $5-8$ and $9-12$. Table~\ref{tab:abl_layer} shows that reconstructions by excluding the gradients of earlier layers are more accurate than those of deeper layers, whereas dropping the later stage alters the recovery the most. In other words, gradients of deeper layers are more informative for inversion - see Fig.~\ref{fig:layerwise}(a) for qualitative comparisons. 

\subsubsection{Attention is All That Reveals}
We next perform two component-wise data leakage studies on the ViT-B/16 model by only utilizing the gradients of MLP or MSA blocks for the inversion attacks. We present results with Table~\ref{tab:abl_block} and Fig.~\ref{fig:layerwise}(b). Table~\ref{tab:abl_block} demonstrates the importance of MSA gradients, as its reconstructions have significantly better image quality than images synthesized by MLP gradients. As illustrated in Fig.~\ref{fig:layerwise}(b), reconstructions from MLP gradients lack important details, whereas utilizing the gradients of MSA layers alone can already yield high-quality reconstructions.     
\input{Images/layer_wise}

\input{tables/abl_layerwise}
\input{tables/abl_blockwise}

%% file: tables/abl_benchmarks.tex
\begin{table}[!t]
\centering
\resizebox{.82\linewidth}{!}{
\begin{tabular}{lccccc}
\toprule
\multirow{2}{*}{\textbf{Network}}  &\multirow{2}{*}{\textbf{Distilled}}& \multicolumn{4}{c}{\textbf{Image Reconstruction Metric}}\\
\cmidrule{3-5}
&  & PSNR $\uparrow$ & FFT$_\text{2D}$ $\downarrow$  & LPIPS $\downarrow$  \\
\midrule
DeiT-T/16~\cite{touvron2021training} &No& $12.243$ & $0.079$ & $0.489$    \\
DeiT-T/16~\cite{touvron2021training} &Yes& $13.212$ & $0.076$ & $0.454$    \\
DeiT-S/16~\cite{touvron2021training}  &No & $12.664$ & $0.059$ & $0.461$   \\
DeiT-S/16~\cite{touvron2021training}  &Yes & $13.092$ & $0.055$ & $0.419$   \\
DeiT-B/16~\cite{touvron2021training}  &No & $13.252$ & $0.058$ & $0.413$   \\
DeiT-B/16~\cite{touvron2021training}  &Yes & $13.708$ & $0.041$ & $0.407$   \\
ViT-T/16~\cite{DBLP:conf/iclr/DosovitskiyB0WZ21} & - & $12.521$ & $0.062$ & $0.483$   \\
ViT-S/32~\cite{DBLP:conf/iclr/DosovitskiyB0WZ21} &- & $12.365$ & $0.063$ & $0.505$   \\
ViT-S/16~\cite{DBLP:conf/iclr/DosovitskiyB0WZ21} &- & $13.658$ & $0.042$ & $0.412$    \\
ViT-B/32~\cite{DBLP:conf/iclr/DosovitskiyB0WZ21} &- & $13.599$ & $0.048$ & $0.436$   \\
ViT-B/16~\cite{DBLP:conf/iclr/DosovitskiyB0WZ21}&- & $\mathbf{15.515}$ & $\mathbf{0.032}$ & $\mathbf{0.295}$ \\
\bottomrule
\end{tabular}}
\caption{Quantitative comparisons of image reconstruction quality from gradient inversion of various ViT and DeiT models on ImageNet1K.}
\label{tab:abl_bench}
\vspace{-1mm}
\end{table}

%% file: Images/large_bs_graph.tex
\begin{figure}
\centering\includegraphics[width=.8\linewidth]{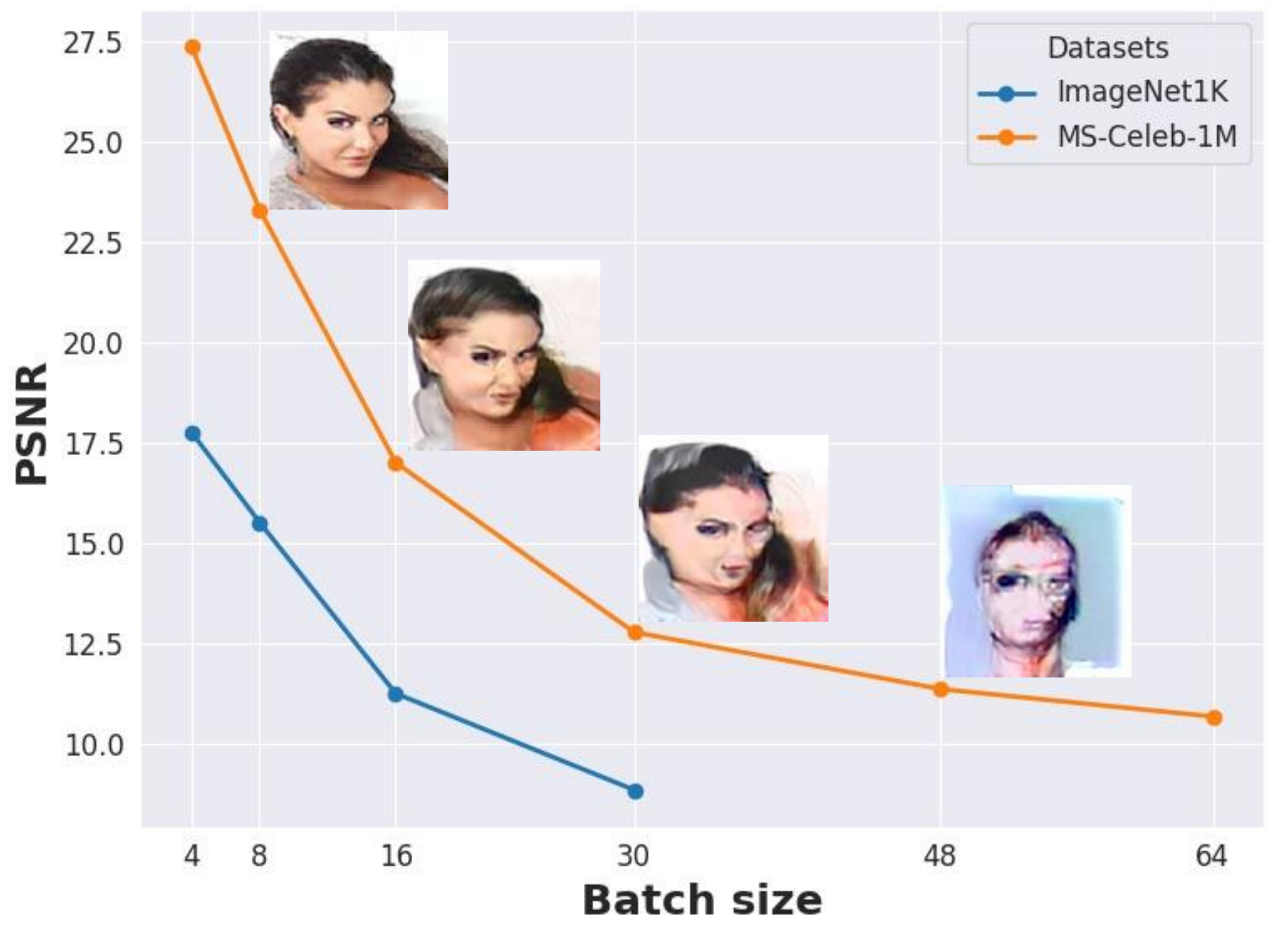}
\caption{Effect of increasing batch size on the quality of image recovery. ImageNet and MS-Celeb-1M images are reconstructed in $224\times224$ px and $112\times112$ px respectively. Representative sample reconstructions are presented for batch sizes of $8,16,30$ and $48$. The maximum number of batch sizes is limited to $30$ for ImageNet dataset amid GPU memory constraint.
}
\label{fig:large_bs_psnr}
\end{figure}

%% file: Images/large_bs.tex
\begin{figure}[t]
\centering
\resizebox{\linewidth}{!}{
\begingroup
\renewcommand*{\arraystretch}{0.3}
\begin{tabular}{c|ccc}
\includegraphics[width=0.3\linewidth,clip,trim=5px 0 0 4px]{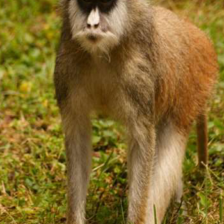} &
\includegraphics[width=0.3\linewidth,clip,trim=5px 0 0 4px]{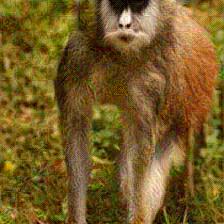} &
\includegraphics[width=0.3\linewidth,clip,trim=5px 0 0 4px]{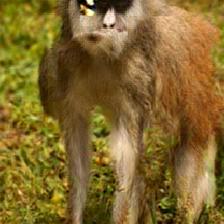} &
\includegraphics[width=0.3\linewidth,clip,trim=5px 0 0 4px]{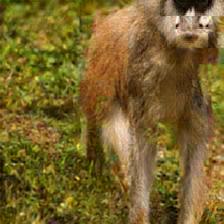}
\\
\multirow{2}{*}{\textbf{Original}} & batch size $8$ & batch size $16$ & batch size $30$ 
\vspace{0.5mm} \\
 & \multicolumn{3}{c}{\textbf{Restored}} \\
\end{tabular}
\endgroup
}
\caption{Visual comparison of reconstruction quality with different batch sizes on ImageNet. Although GradViT recovers major visual features, the quality decreases with increasing batch size.} 
\vspace{-3mm}
\label{fig:vanishing_feature}
\end{figure}

%% file: Images/layer_wise.tex
\begin{figure}[h]
\centering

\resizebox{0.99\linewidth}{!}{

\begingroup
\renewcommand*{\arraystretch}{0.3}
\begin{tabular}{c|ccc}
\textbf{Original} & \multicolumn{3}{c}{\textbf{Recovery}} \\
                  & \multicolumn{3}{c}{(w/o full grad., \textit{\textbf{layer-wise}} distinction)} \\
\includegraphics[width=0.3\linewidth,clip,trim=5px 0 0 4px]{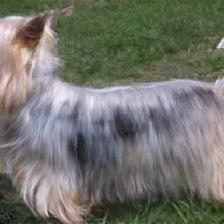} &
\includegraphics[width=0.3\linewidth,clip,trim=5px 0 0 4px]{Images/block_wise/img3_layer0.jpg} &
\includegraphics[width=0.3\linewidth,clip,trim=5px 0 0 4px]{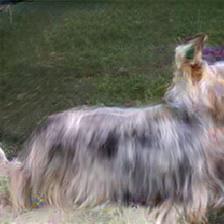} &
\includegraphics[width=0.3\linewidth,clip,trim=5px 0 0 4px]{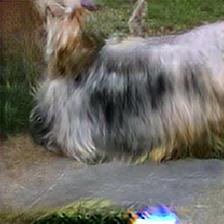}
\\
 & w/o layers $1-4$ & w/o layers $5-8$ & w/o layers $9-12$ \\
& & & \\
& \multicolumn{3}{c}{(a)} \\
& & & \\
& & & \\
& & & \\
& & & \\
\multicolumn{1}{c|}{
\begin{tabular}{c}
\textbf{Recovery} \\
(w/ full grad.) \\
\includegraphics[width=0.3\linewidth,clip,trim=5px 0 0 4px]{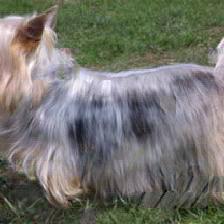} \\ 
\\
\end{tabular}
} &
\multicolumn{3}{c}{
\begin{tabular}{cc}
\multicolumn{2}{c}{\textbf{Recovery}} \\
\multicolumn{2}{c}{(w/o full grad., \textit{\textbf{component-wise}} distinction)} \\
\includegraphics[width=0.3\linewidth,clip,trim=5px 0 0 4px]{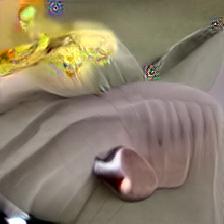} &
\includegraphics[width=0.3\linewidth,clip,trim=5px 0 0 4px]{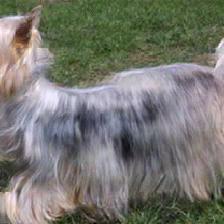} \\
w/ MLP grad., w/o others & w/ attn. grad., w/o others \\
& \\
\multicolumn{2}{c}{(b)} 
\end{tabular}
}  \\
\end{tabular}
\endgroup
}

\caption{Reconstructed images from layer-wise and component-wise ablation studies using a batch size of 8. Later layers (9-12) contain the most critical information that leads to data leakage. The component-wise studies show that gradients of MSA blocks have more critical information than those of MLP blocks. See supplementary materials for more visualizations.
}
\vspace{-0.5mm}
\label{fig:layerwise}
\end{figure}

%% file: tables/abl_layerwise.tex
\begin{table}[t]
\centering
\resizebox{.78\linewidth}{!}{
\begin{tabular}{lcccc}
\toprule
\multirow{2}{*}{\textbf{Layer-wise Gradients}}  & \multicolumn{4}{c}{\textbf{Image Reconstruction Metric}}\\
\cmidrule{2-5}
&PSNR $\uparrow$& FFT$_\text{2D}$ $\downarrow$  & LPIPS $\downarrow$  \\
\midrule
All (baseline)  & $15.515$ & $0.032$  & $0.295$ \\
\midrule
w/o Layers 1-4  & $13.982$ & $0.047$  & $0.412$ \\
w/o Layers 5-8  & $11.086$ & $0.086$  & $0.555$ \\
w/o Layers 9-12  & $10.284$ & $0.091$  & $0.598$ \\
\bottomrule
\end{tabular}}
\caption{Effect of layer-wise gradients on ViT-B/16 reconstructions.}
\label{tab:abl_layer}
\vspace{-1mm}
\end{table}

%% file: tables/abl_blockwise.tex
\begin{table}[t]
\centering
\resizebox{.84\linewidth}{!}{
\begin{tabular}{lcccc}
\toprule
\multirow{2}{*}{\textbf{Component-wise Gradients}}  & \multicolumn{4}{c}{\textbf{Image Reconstruction Metric}}\\
\cmidrule{2-5}
&PSNR $\uparrow$& FFT$_\text{2D}$ $\downarrow$  & LPIPS $\downarrow$  \\
\midrule
All (baseline)  & $15.515$ & $0.032$  & $0.295$ \\
\midrule
w/ MLP, w/o others  & $12.256$ & $0.066$  & $0.568$ \\
w/ MSA, w/o others  & $13.559$ & $0.047$  & $0.408$ \\
\bottomrule
\end{tabular}}
\caption{Effect of component-wise gradients on ViT-B/16 reconstructions.}
\label{tab:abl_block}
\vspace{-1mm}
\end{table}

%% file: sec8_supp_arxiv.tex
\section*{Appendix}
\appendix
\renewcommand{\thesection}{\Alph{section}}
\renewcommand\thefigure{S.\arabic{figure}}
\setcounter{figure}{0}
\renewcommand\thetable{S.\arabic{table}}
\setcounter{table}{0}
We provide more experimental details in the following
sections. First, we elaborate on person identification analysis to evaluate inversion strength in Section~\ref{sec:supp_person_id}. Then, we demonstrate efficacy of the proposed loss scheduling technique in Section~\ref{sec:supp_loss_scheduler}. We include additional supplementary images for ablation studies in Section~\ref{sec:supp_ablations}, and more visual examples for GradViT in Section~\ref{sec:supp_more_samples}.

\vspace{-2mm}
\section{Person Identification}
\label{sec:supp_person_id}
\begin{figure*}[t]
\centering

\resizebox{\linewidth}{!}{
\begingroup
\renewcommand*{\arraystretch}{0.3}
\begin{tabular}{c}

\includegraphics[width=1.1\linewidth]{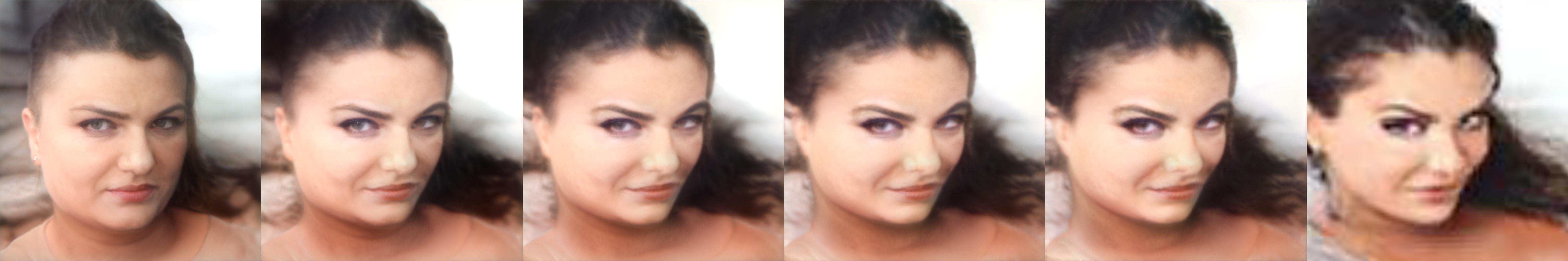} \\
 \small{Latent optimization outputs - recovered input from batch size 8}\\ 
    \vspace{1mm} \\

 \includegraphics[width=1.1\linewidth]{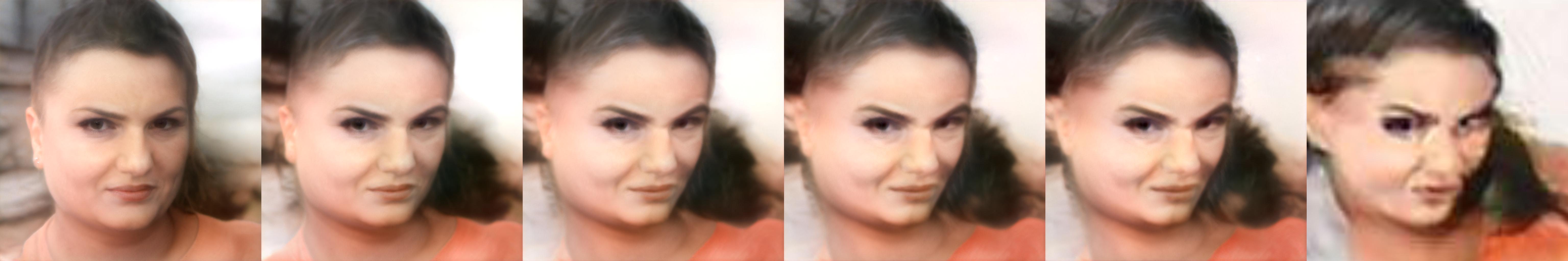} \\
 \small{Latent optimization outputs - recovered input from batch size 16}
 
 \vspace{1mm}\\

\includegraphics[width=1.1\linewidth]{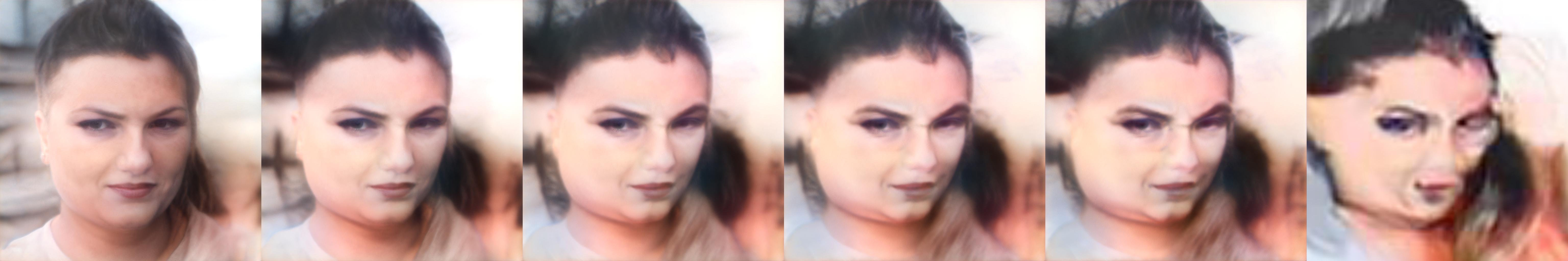} \\
 \small{Latent optimization outputs - recovered input from batch size 32}
 \vspace{-1mm}\\ 

\end{tabular}
\endgroup
}
\caption{Step-by-step latent optimization of GradViT outputs. Recovered image shown on the right.}
\label{fig:style_gan_iter}
\end{figure*}

In this section, we study the likelihood of person identification as a function of the batch size, and also leverage a StyleGAN2~\cite{karras2020analyzing} network to improve the image fidelity. Specifically, we utilize an iterative refinement approach~\cite{alaluf2021restyle} based on a pre-trained StyleGAN2 for latent space optimization and finding the closest real image. Fig.~\ref{fig:style_gan_iter} illustrates the outputs of the latent optimization which uses a GradViT recovered image as an input. 

To quantify person identification, we utilize the Image Identifiablity Precision (IIP) as studied by Yin \etal~\cite{yin2021see} to check on the level of data leakage across varying batch sizes. Specifically, we consider a total number of $15000$ distinct subjects randomly selected from the MS-Celeb-1M dataset. 
The experiments are performed once for reconstructions of a given batch size. For IIP calculation, we extract deep feature embeddings using an ImageNet-1K pre-trained ResNet-50. To compute exact matches, we use k-nearest neighbor clustering to sort the closest training images to the reconstructions in the embedding space. The IIP score is computed as the ratio of number of exact matches to the batch size. 

\begin{figure}[t]
\centering

\resizebox{\linewidth}{!}{
\begingroup
\begin{tabular}{c}

\includegraphics[width=.8\linewidth]{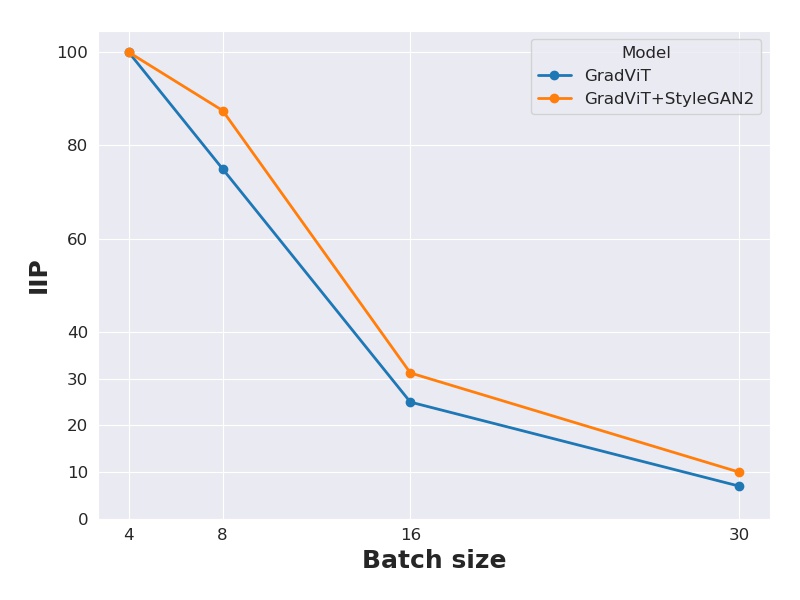} \\
\end{tabular}
\endgroup
}
 \vspace{-4mm}\\
\caption{Effect of batch size on IIP score for recovered images from MS-Celeb-1M dataset. Random guess probability is $0.007\%$ as a reference.}
\label{fig:iip_batch}
\end{figure}

We use the outputs of GradViT and GradViT followed by StyleGAN2 for all person identification experiments (See Fig.~\ref{fig:iip_batch}). We observe that for a batch size of $4$, both models can accurately identify subjects with an IIP score of $100\%$. For a batch size of $8$, GradViT and GradViT+StyleGAN2 yield IIP scores of $75\%$ and $87.5\%$ respectively. We observe that enhancements by latent code optimization result in improved facial recovery and hence increased IIP scores. IIP gradually decrease for both cases amid more gradient averaging at larger batch sizes.

\begin{figure}[t]
\centering

\resizebox{\linewidth}{!}{
\begingroup
\begin{tabular}{c}

\includegraphics[width=.8\linewidth]{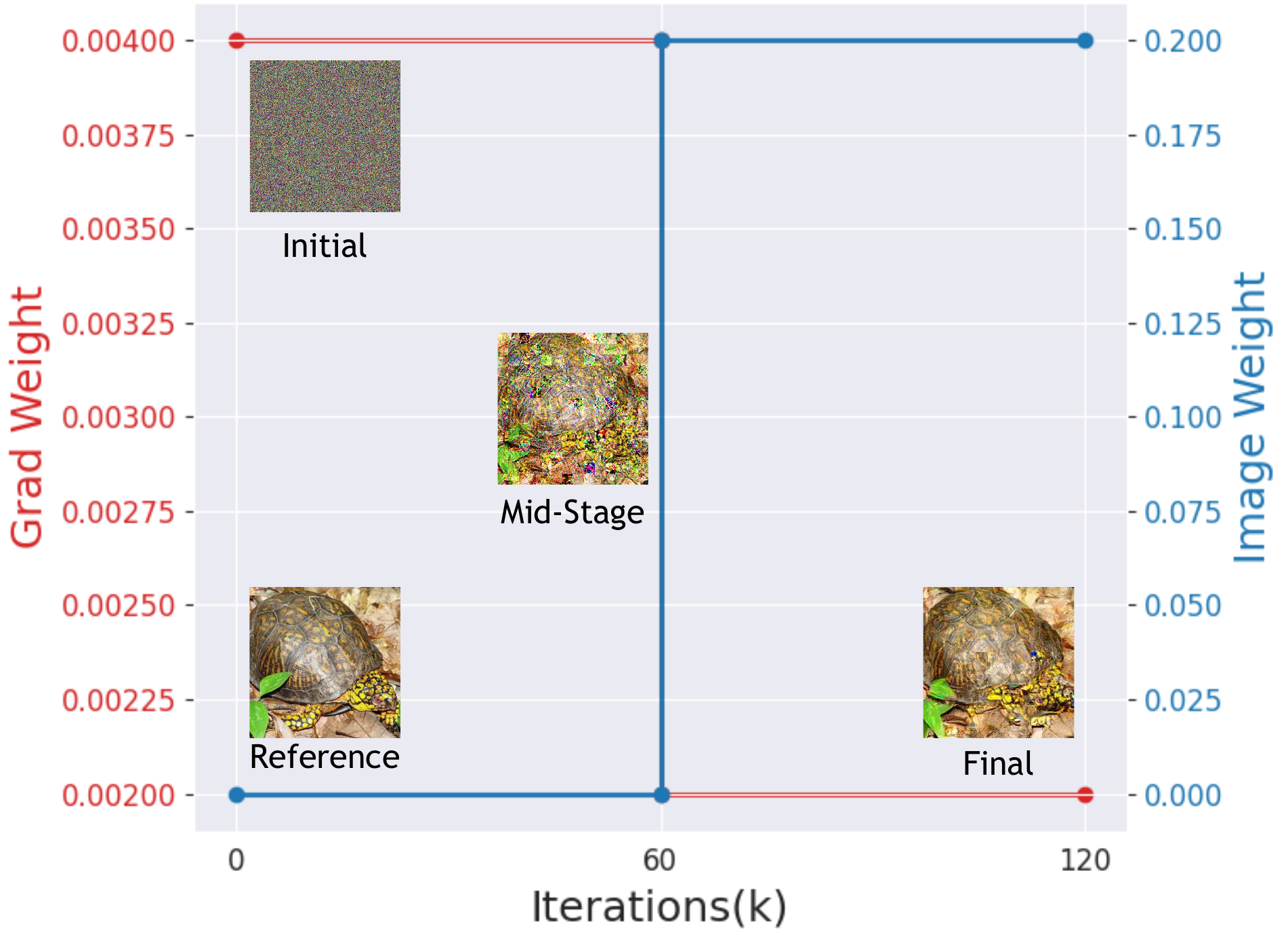} \\
\end{tabular}
\endgroup
}
\vspace{-4mm}\\
\caption{Effect of loss scheduler on optimization progression of reconstructions.}
\label{fig:loss_schedule}
\end{figure}


\begin{figure}[h]
\centering

\resizebox{0.95\linewidth}{!}{

\begingroup
\renewcommand*{\arraystretch}{0.3}
\begin{tabular}{c|ccc}
\textbf{Original} & \multicolumn{3}{c}{\textbf{Recovery}} \\
                  & \multicolumn{3}{c}{(w/o full grad., \textit{\textbf{layer-wise}} distinction)} \\
\includegraphics[width=0.3\linewidth,clip,trim=5px 0 0 4px]{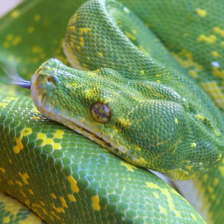} &
\includegraphics[width=0.3\linewidth,clip,trim=5px 0 0 4px]{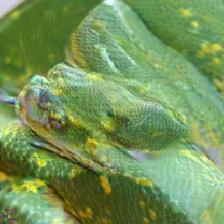} &
\includegraphics[width=0.3\linewidth,clip,trim=5px 0 0 4px]{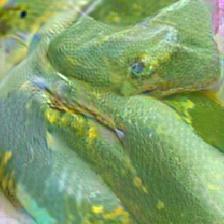} &
\includegraphics[width=0.3\linewidth,clip,trim=5px 0 0 4px]{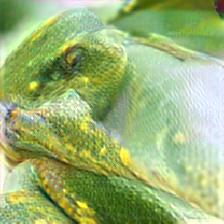}
\\
\includegraphics[width=0.3\linewidth,clip,trim=5px 0 0 4px]{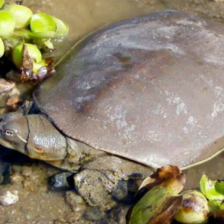} &
\includegraphics[width=0.3\linewidth,clip,trim=5px 0 0 4px]{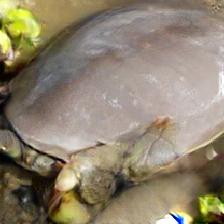} &
\includegraphics[width=0.3\linewidth,clip,trim=5px 0 0 4px]{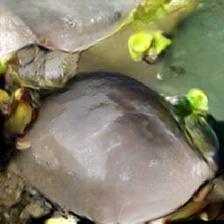} &
\includegraphics[width=0.3\linewidth,clip,trim=5px 0 0 4px]{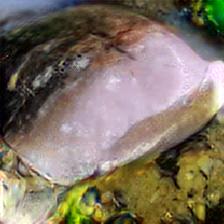}
\\
\includegraphics[width=0.3\linewidth,clip,trim=5px 0 0 4px]{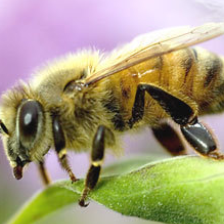} &
\includegraphics[width=0.3\linewidth,clip,trim=5px 0 0 4px]{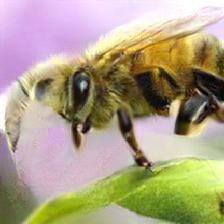} &
\includegraphics[width=0.3\linewidth,clip,trim=5px 0 0 4px]{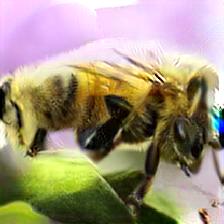} &
\includegraphics[width=0.3\linewidth,clip,trim=5px 0 0 4px]{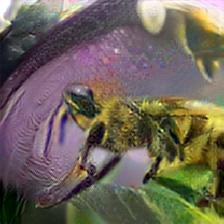}
\\
\includegraphics[width=0.3\linewidth,clip,trim=5px 0 0 4px]{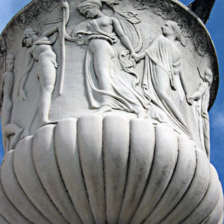} &
\includegraphics[width=0.3\linewidth,clip,trim=5px 0 0 4px]{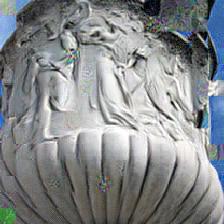} &
\includegraphics[width=0.3\linewidth,clip,trim=5px 0 0 4px]{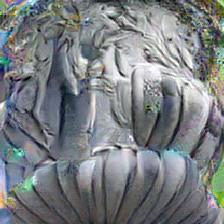} &
\includegraphics[width=0.3\linewidth,clip,trim=5px 0 0 4px]{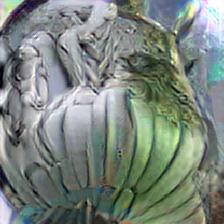}
\\
\includegraphics[width=0.3\linewidth,clip,trim=5px 0 0 4px]{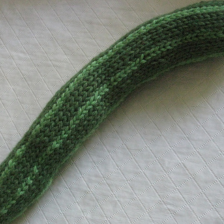} &
\includegraphics[width=0.3\linewidth,clip,trim=5px 0 0 4px]{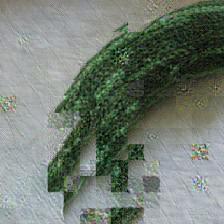} &
\includegraphics[width=0.3\linewidth,clip,trim=5px 0 0 4px]{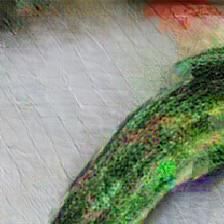} &
\includegraphics[width=0.3\linewidth,clip,trim=5px 0 0 4px]{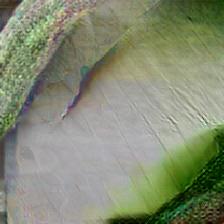}
\\
 & w/o layers $1-4$ & w/o layers $5-8$ & w/o layers $9-12$ \\
& & & \\
\end{tabular}
\endgroup
}
\caption{Reconstructed images from layer-wise ablation.
}
\vspace{-1mm}
\label{fig:layer_wise_supp}
\end{figure}

\begin{figure}[h]
\centering

\resizebox{0.95\linewidth}{!}{

\begingroup
\renewcommand*{\arraystretch}{0.3}
\begin{tabular}{c|ccc}
\textbf{Original} & \multicolumn{3}{c}{\textbf{Recovery}} \\
                  & \multicolumn{3}{c}{(w/o full grad., \textit{\textbf{component-wise}} distinction)} \\
\includegraphics[width=0.3\linewidth,clip,trim=5px 0 0 4px]{Images/block_wise/img2_orig.png} &
\includegraphics[width=0.3\linewidth,clip,trim=5px 0 0 4px]{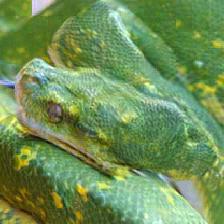} &
\includegraphics[width=0.3\linewidth,clip,trim=5px 0 0 4px]{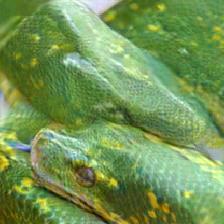} &
\includegraphics[width=0.3\linewidth,clip,trim=5px 0 0 4px]{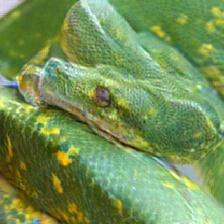}
\\
\includegraphics[width=0.3\linewidth,clip,trim=5px 0 0 4px]{Images/block_wise/img1_orig.png} &
\includegraphics[width=0.3\linewidth,clip,trim=5px 0 0 4px]{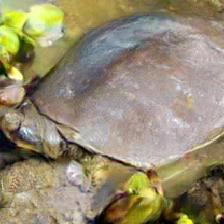} &
\includegraphics[width=0.3\linewidth,clip,trim=5px 0 0 4px]{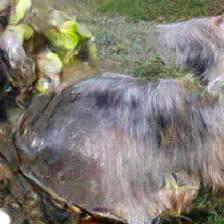} &
\includegraphics[width=0.3\linewidth,clip,trim=5px 0 0 4px]{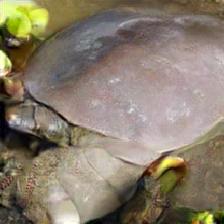}
\\
\includegraphics[width=0.3\linewidth,clip,trim=5px 0 0 4px]{Images/block_wise/img4_orig.png} &
\includegraphics[width=0.3\linewidth,clip,trim=5px 0 0 4px]{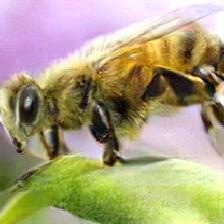} &
\includegraphics[width=0.3\linewidth,clip,trim=5px 0 0 4px]{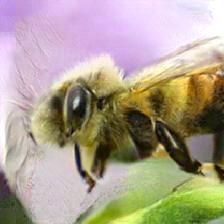} &
\includegraphics[width=0.3\linewidth,clip,trim=5px 0 0 4px]{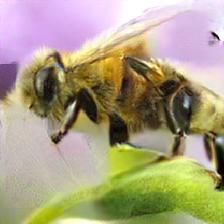}
\\
\includegraphics[width=0.3\linewidth,clip,trim=5px 0 0 4px]{Images/block_wise/img5_orig.png} &
\includegraphics[width=0.3\linewidth,clip,trim=5px 0 0 4px]{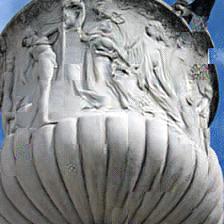} &
\includegraphics[width=0.3\linewidth,clip,trim=5px 0 0 4px]{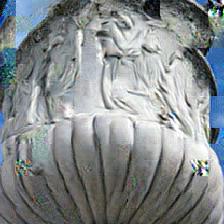} &
\includegraphics[width=0.3\linewidth,clip,trim=5px 0 0 4px]{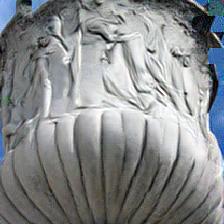}
\\\includegraphics[width=0.3\linewidth,clip,trim=5px 0 0 4px]{Images/block_wise/img6_orig.png} &
\includegraphics[width=0.3\linewidth,clip,trim=5px 0 0 4px]{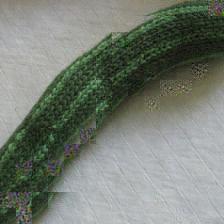} &
\includegraphics[width=0.3\linewidth,clip,trim=5px 0 0 4px]{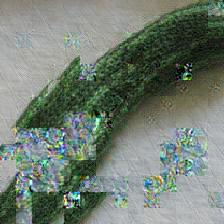} &
\includegraphics[width=0.3\linewidth,clip,trim=5px 0 0 4px]{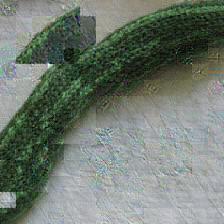}
\\
 & w/ full grad. & w/ MLP grad. & w/ attn. grad. \\
& & & \\
\end{tabular}
\endgroup
}

\caption{Reconstructed images from component-wise ablation.
}
\vspace{-1mm}
\label{fig:block_wise_supp}
\end{figure}


\label{sec:grad_recon}

\vspace{-2mm}
\section{Loss Scheduler}
\label{sec:supp_loss_scheduler}
In Fig.~\ref{fig:loss_schedule}, we demonstrate the effect of our proposed loss scheduler on balancing the training between the gradient matching loss and the image prior. As observed by the progression of optimization from random noise to the final image, gradient matching phase obtains most of the semantics in the image by the mid-training. However, the recovered image lacks detailed information and suffers from low fidelity. By enabling the image prior loss and decreasing the contribution of gradient matching, the visual realism is significantly improved and more fine-grained detailed are recovered. 
\vspace{-2mm}
\section{Ablation Studies Additional Examples}
\label{sec:supp_ablations}
We provide additional qualitative visualizations of data leakage analysis by layer-wise (Fig.~\ref{fig:layer_wise_supp}) and component-wise (Fig.~\ref{fig:block_wise_supp}) analysis of ViT/B-16~\cite{DBLP:conf/iclr/DosovitskiyB0WZ21} architecture. Inversion results obtained from batch of 8 images.
\vspace{-2mm}
\section{Limitations}
\label{sec:impact}
GradViT remains computationally intensive. However, this offers security benefits in reality, given that the associated computation burden may hinder gradient inversion at scale.

\vspace{-2mm}
\section{More Inversion Examples}
\label{sec:supp_more_samples}
For ImageNet1K dataset, Fig.~\ref{fig:grad_invert_samples} depicts additional recovered images from gradient inversion of vision transformers using GradViT for varying batch sizes. In addition, Fig.~\ref{fig:grad_invert_samples_face_supp} demonstrates additional reconstructions from gradient inversion of FaceTransformer~\cite{zhong2021face} model using GradViT for different batch sizes.

\input{Images/fig_supp_imgnet_recon}
\input{Images/fig_supp_imgnet_recon_face}
\section{Face Reconstruction Quantitative Analysis}
\label{sec:face_comp}
Table.~\ref{tab:abl_face_batch} presents the quantitative benchmarks of image reconstruction quality for gradient inversion from batch sizes of 4 and 8 using images in MS-Celeb-1M dataset. As expected, for all image reconstruction metrics, the reconstruction quality decreases with increasing batch size. 
\input{tables/face_comparisons}


%% file: Images/fig_supp_imgnet_recon.tex
\noindent\begin{figure*}[!t]
\centering

\resizebox{.99\linewidth}{!}{
\begingroup
\renewcommand*{\arraystretch}{0.3}
\begin{tabular}{ccc}
\includegraphics[width=0.33\linewidth,clip,trim=5px 0 0 4px]{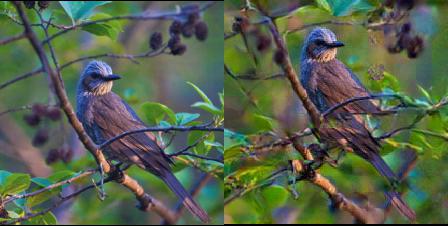} &
\includegraphics[width=0.33\linewidth,clip,trim=5px 0 0 4px]{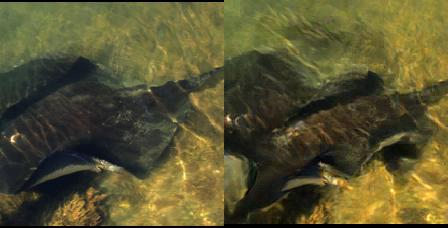} &
\includegraphics[width=0.33\linewidth,clip,trim=5px 0 0 4px]{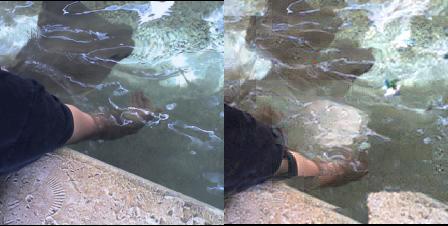}
\\
\includegraphics[width=0.33\linewidth,clip,trim=5px 0 0 4px]{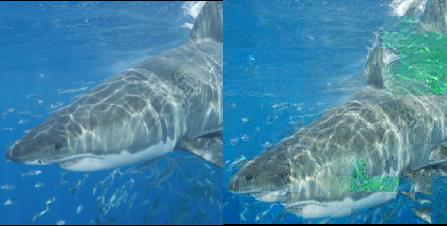} &
\includegraphics[width=0.33\linewidth,clip,trim=5px 0 0 4px]{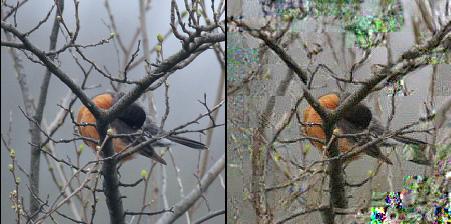} &
\includegraphics[width=0.33\linewidth,clip,trim=5px 0 0 4px]{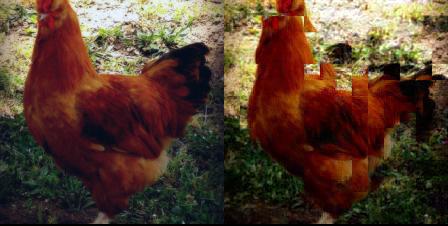}
\\
\includegraphics[width=0.33\linewidth,clip,trim=5px 0 0 4px]{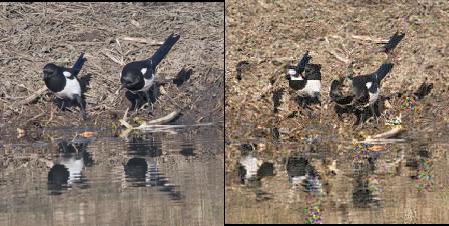} &
\includegraphics[width=0.33\linewidth,clip,trim=5px 0 0 4px]{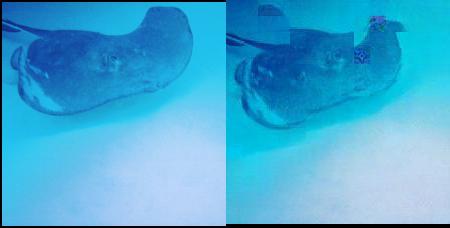} &
\includegraphics[width=0.33\linewidth,clip,trim=5px 0 0 4px]{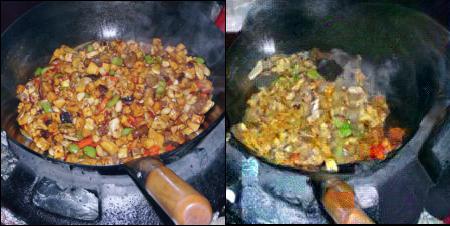} \\
\multicolumn{3}{c}{batch size $8$}
\end{tabular}
\endgroup
}

\resizebox{.99\linewidth}{!}{
\begingroup
\renewcommand*{\arraystretch}{0.3}
\begin{tabular}{ccc}
\includegraphics[width=0.33\linewidth,clip,trim=5px 0 0 4px]{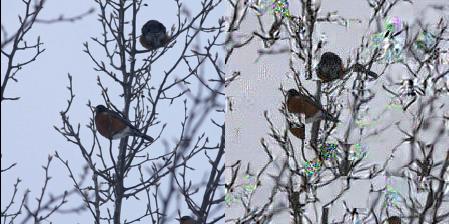} &
\includegraphics[width=0.33\linewidth,clip,trim=5px 0 0 4px]{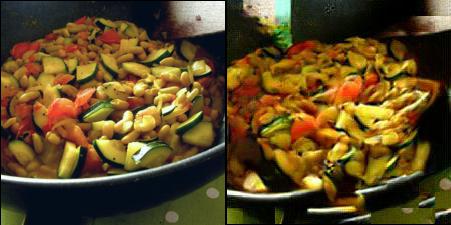}  &
\includegraphics[width=0.33\linewidth,clip,trim=5px 0 0 4px]{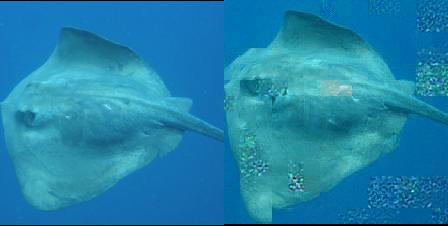}  
\\
\includegraphics[width=0.33\linewidth,clip,trim=5px 0 0 4px]{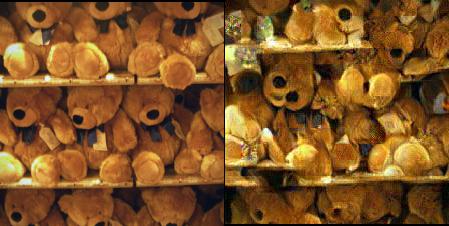}  &
\includegraphics[width=0.33\linewidth,clip,trim=5px 0 0 4px]{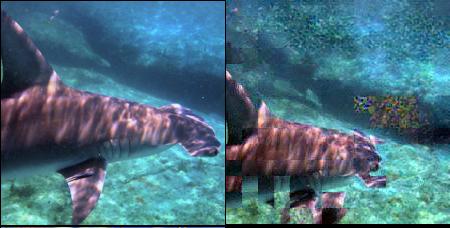}  &
\includegraphics[width=0.33\linewidth,clip,trim=5px 0 0 4px]{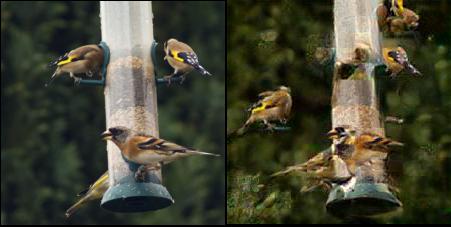}  
\\
\includegraphics[width=0.33\linewidth,clip,trim=5px 0 0 4px]{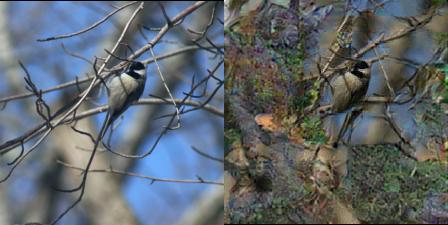}  &
\includegraphics[width=0.33\linewidth,clip,trim=5px 0 0 4px]{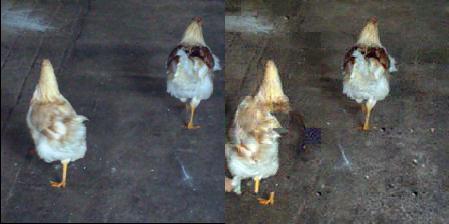}  &
\includegraphics[width=0.33\linewidth,clip,trim=5px 0 0 4px]{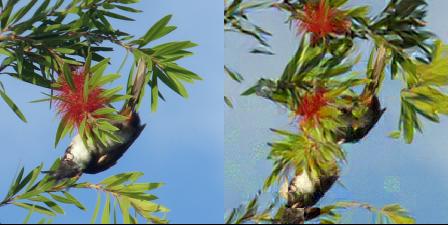}  
\\
\multicolumn{3}{c}{batch size $16$}
\end{tabular}
\endgroup
}

\caption{Inverting ViT-B/16 gradients on the ImageNet-1K validation set. Pair of (left) original sample and its (right) recovery. }
\label{fig:grad_invert_samples}
\end{figure*}

%% file: Images/fig_supp_imgnet_recon_face.tex
\noindent\begin{figure*}[]
\centering
\resizebox{.92\linewidth}{!}{
\begingroup
\renewcommand*{\arraystretch}{0.3}
\begin{tabular}{ccc}
\includegraphics[width=0.33\linewidth,clip,trim=5px 0 0 4px]{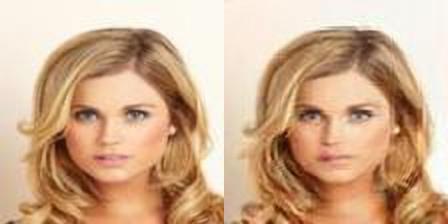} &
\includegraphics[width=0.33\linewidth,clip,trim=5px 0 0 4px]{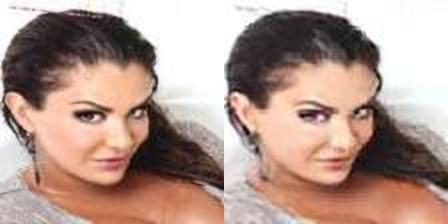} &
\includegraphics[width=0.33\linewidth,clip,trim=5px 0 0 4px]{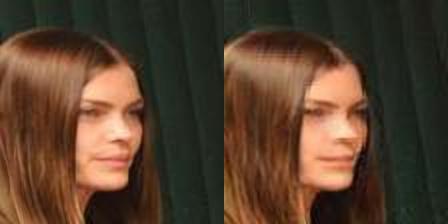}
\\
\includegraphics[width=0.33\linewidth,clip,trim=5px 0 0 4px]{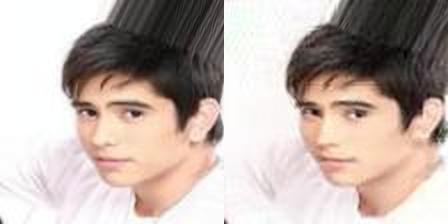} &
\includegraphics[width=0.33\linewidth,clip,trim=5px 0 0 4px]{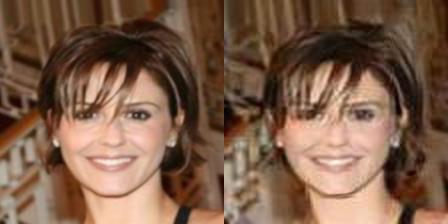} &
\includegraphics[width=0.33\linewidth,clip,trim=5px 0 0 4px]{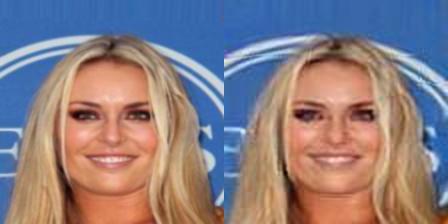}
\\
\includegraphics[width=0.33\linewidth,clip,trim=5px 0 0 4px]{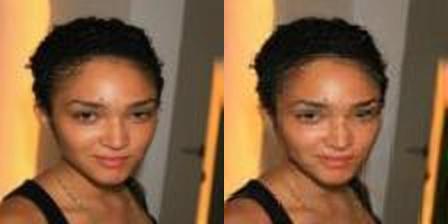} &
\includegraphics[width=0.33\linewidth,clip,trim=5px 0 0 4px]{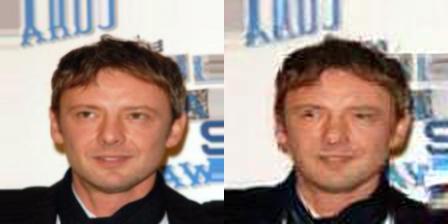} &
\includegraphics[width=0.33\linewidth,clip,trim=5px 0 0 4px]{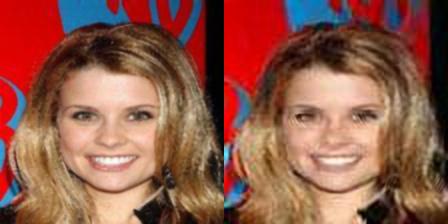} \\
\multicolumn{3}{c}{batch size $4$}
\vspace{1mm}
\end{tabular}
\endgroup
}

\resizebox{.92\linewidth}{!}{
\begingroup
\renewcommand*{\arraystretch}{0.3}
\begin{tabular}{ccc}
\includegraphics[width=0.33\linewidth,clip,trim=5px 0 0 4px]{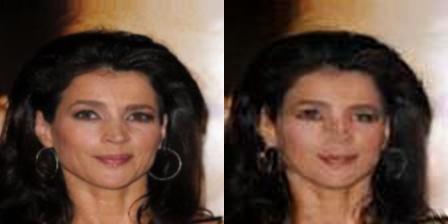} &
\includegraphics[width=0.33\linewidth,clip,trim=5px 0 0 4px]{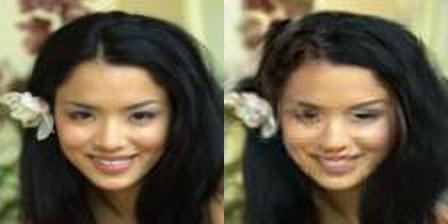}  &
\includegraphics[width=0.33\linewidth,clip,trim=5px 0 0 4px]{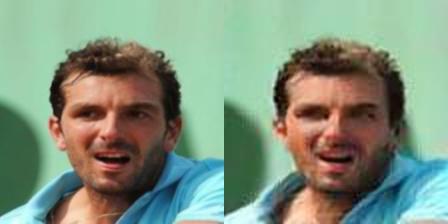}  
\\
\includegraphics[width=0.33\linewidth,clip,trim=5px 0 0 4px]{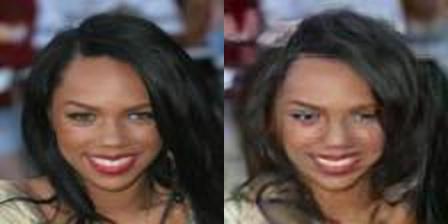}  &
\includegraphics[width=0.33\linewidth,clip,trim=5px 0 0 4px]{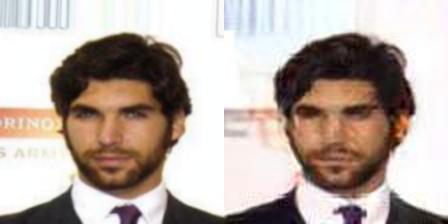}  &
\includegraphics[width=0.33\linewidth,clip,trim=5px 0 0 4px]{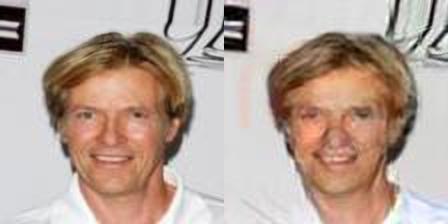}  
\\
\multicolumn{3}{c}{batch size $8$}
\vspace{1mm}
\end{tabular}
\endgroup
}

\caption{Additional examples of information leakage when inverting FaceTransformer gradients on the MS-Celeb-1M validation set. Each block containing a pair of (left) original sample and its (right) reconstruction by GradViT. }
\label{fig:grad_invert_samples_face_supp}
\end{figure*}

%% file: tables/face_comparisons.tex
\begin{table}[!t]
\centering
\resizebox{.6\linewidth}{!}{
\begin{tabular}{lcccc}
\toprule
\multirow{2}{*}{\textbf{Batch Size}}  & \multicolumn{4}{c}{\textbf{Image Reconstruction Metric}}\\
\cmidrule{2-5}
&PSNR $\uparrow$& FFT$_\text{2D}$ $\downarrow$  & LPIPS $\downarrow$  \\
\midrule
4  & $27.370$ & $0.001$  & $0.030$ \\
8 & $23.313$ & $0.008$  & $0.101$ \\
\bottomrule
\end{tabular}}
\caption{Quantitative benchmarks of image reconstruction quality from batch sizes of 4 and 8 images in MS-Celeb-1M dataset.}
\label{tab:abl_face_batch}
\vspace{-1mm}
\end{table}